\documentclass{article}

\usepackage{microtype}
\usepackage{graphicx}
\usepackage{subfigure}
\usepackage{booktabs} 

\usepackage{hyperref}

\usepackage[accepted]{icml2024}

\usepackage{amsmath}
\usepackage{amssymb}
\usepackage{mathtools}
\usepackage{amsthm}

\usepackage[capitalize,noabbrev]{cleveref}

\theoremstyle{plain}
\newtheorem{theorem}{Theorem}[section]

\newtheorem{corollary}[theorem]{Corollary}
\theoremstyle{definition}

\newtheorem{assumption}[theorem]{Assumption}
\theoremstyle{remark}

\usepackage[textsize=tiny]{todonotes}

\icmltitlerunning{SignSGD with Federated Defense: Harnessing Adversarial Attacks through Gradient Sign Decoding}

\begin{document}

\twocolumn[

\icmltitle{SignSGD with Federated Defense: \\ Harnessing Adversarial Attacks through Gradient Sign Decoding}



\icmlsetsymbol{equal}{*}

\begin{icmlauthorlist}
\icmlauthor{Chanho Park}{postech}
\icmlauthor{Namyoon Lee}{korea}
\end{icmlauthorlist}

\icmlaffiliation{postech}{Department of Electrical Engineering, Pohang University of Science and Technology, Pohang, South Korea}
\icmlaffiliation{korea}{School of Electrical Engineering, Korea University, Seoul, South Korea}

\icmlcorrespondingauthor{Chanho Park}{chanho26@postech.ac.kr}
\icmlcorrespondingauthor{Namyoon Lee}{namyoon@korea.ac.kr}

\icmlkeywords{Distributed learning, federated learning, adversarial attacks, binary symmetric channel, convergence rate}

\vskip 0.3in
]



\printAffiliationsAndNotice{}  

\begin{abstract}

Distributed learning is an effective approach to accelerate model training using multiple workers. However, substantial communication delays emerge between workers and a parameter server due to massive costs associated with communicating gradients. SignSGD with majority voting (signSGD-MV) is a simple yet effective optimizer that reduces communication costs through one-bit quantization, yet the convergence rates considerably decrease as adversarial workers increase. In this paper, we show that the convergence rate is invariant as the number of adversarial workers increases, provided that the number of adversarial workers is smaller than that of benign workers. The key idea showing this counter-intuitive result is our novel signSGD with federated defense (signSGD-FD). Unlike the traditional approaches, signSGD-FD exploits the gradient information sent by adversarial workers with the proper weights, which are obtained through gradient sign decoding. Experimental results demonstrate signSGD-FD achieves superior convergence rates over traditional algorithms in various adversarial attack scenarios.

\end{abstract}

\section{Introduction} \label{sec:intro}

Distributed stochastic gradient descent (SGD) stands as a widely adopted technique for tackling large-scale optimization challenges involving data parallelism \cite{bottou2010large, dean2012large}. Theoretically, synchronous distributed SGD has the potential to significantly boost the training speed of high-dimensional models in proportion to the number of workers. Nonetheless, the practical implementation of such distributed SGD encounters hurdles, notably the substantial communication costs associated with exchanging gradient information between the central server and the individual workers. This communication expense scales linearly with the number of workers. For instance, the cutting-edge large language model model \cite{zhao2023survey}, using a few billion parameters, necessitates an exchange of tens of Giga bytes of information per iteration for both each worker and the central server. This renders it impractical for distributed training, particularly in environments with limited communication networks. To resolve this communication bottleneck issue, it becomes imperative to devise communication-efficient distributed learning algorithms that can mitigate communication costs while upholding high learning performance.



In recent years, various techniques have been proposed with the goal of effectively reducing the communication load in distributed learning \cite{kairouz2021advances}. The primary approach to cost reduction involves edge devices performing lossy compression on locally computed gradient, which are then transmitted to the parameter server. One simple yet effective algorithm is signSGD with majority voting (signSGD-MV) \cite{bernstein2018asignsgd}, in which each worker quantizes the locally computed stochastic gradients with their signs and sends them to the server. Then, the server aggregates the one-bit gradient information using the MV principle and shares the aggregated one with the workers for performing the model update. 

Adversarial attacks aim to interfere with the training models in distributed machine learning systems, introducing security vulnerabilities in their predictive outcomes \cite{lyu2020threats, baruch2019little, xie2020fall,blanchard2017machine, alistarh2018byzantine}. For instance, malevolent workers seek to compromise the model by introducing inaccurate data or manipulating the model's parameters or gradients. SignSGD-MV has also shown to effective to optimize the model against adversarial attack thanks to the majority voting principle \cite{bernstein2018bsignsgd}. Nevertheless, the test accuracy of signSGD-MV deteriorates rapidly as the number of adversarial workers increases. 

\subsection{Contribution}

In this paper, we theoretically and empirically prove that the performance of distributed learning remains unaffected by the number of adversarial workers, as long as the number of legitimate workers exceeds that of adversarial workers.  

\begin{itemize}

  \item  The key idea showing our counter-intuitive result is a novel distributed learning algorithm called signSGD with \textit{federated defense} (signSGD-FD). Diverging from the traditional majority voting approach, federated defense astutely utilizes gradient information derived not only from legitimate workers but also from adversarial workers during aggregation. To elucidate this concept, we offer a coding-theoretical interpretation of signSGD-MV. Building upon this novel interpretation, we introduce a progressive weighted majority voting (WMV) method that dynamically adjusts weights throughout iterations. During each iteration, the server estimates weights by comparing signs between the aggregated gradient through WMV and the local gradient transmitted by workers. This weight estimation process not only aids in identifying adversarial workers but also leverages the weights to enhance resilience against adversarial attacks.

   \item  We present an unified convergence rate analysis for the signSGD-style algorithm incorporating an arbitrary binary aggregation function. Specifically, when employing signSGD-FD, we demonstrate that the convergence rate remains unaffected in the face of adversarial attacks, as long as the number of adversarial workers is less than that of legitimate workers. This finding diverges from the previous convergence rate observed with signSGD-MV, where the convergence rate diminishes with an increasing number of adversarial workers.



    \item  We also provide experimental results on MNIST, CIFAR-10 and CIFAR-100 datasets to validate the robustness of signSGD-FD in the presence of malicious attacks. Compared to signSGD-MV and its variants, signSGD-FD can achieve much higher test accuracy in the presence of stochastic sign flip attacks, especially on the $r=1$ case. Furthermore, we validate the communication efficiency of signSGD-FD by evaluating the communication costs compared to other full-precision attack-robust algorithms.
\end{itemize}

\subsection{Related Works}
\label{sub:related}

{\noindent \bf Gradient compression:} Gradient compression techniques can be categorized into quantization, which compresses the gradient vector into a limited set of codewords, and sparsification, which selectively updates a small number of gradient coordinates to optimize models. Noteworthy quantization methods encompass \cite{seide20141, alistarh2017qsgd, bernstein2018asignsgd, gandikota2021vqsgd, honig2022dadaquant}, while sparsification techniques include \cite{aji2017sparse, wangni2018gradient, stich2018sparsified, rothchild2020fetchsgd, li2022near}. To significantly reduce communication costs, some approaches, as exemplified by \cite{wen2017terngrad, basu2019qsparse, sattler2019robust, park2023sparse, li2023analysis}, integrate both quantization and sparsification. Various adaptations of the signSGD-MV algorithm, such as those proposed in \cite{karimireddy2019error, zheng2019communication, jin2020stochastic, sun2023momentum, jin2024sign}, have been introduced to address additional practical challenges.

{\noindent \bf Robustness to adversarial attacks:} A decentralized learning system is susceptible to malicious attacks, as adversarial attackers can engage in the system. A common attack method in distributed learning is the Byzantine attack \cite{lamport2019byzantine}. To counteract such threats, various defense algorithms modify the aggregation process, which traditionally involves computing the average of workers' gradients. These algorithms include coordinate-wise median \cite{yin2018byzantine}, geometric median \cite{blanchard2017machine, guerraoui2018hidden}, center clipping \cite{karimireddy2021learning}, and weighted aggregation \cite{pillutla2022robust}. The signSGD-MV algorithm is recognized as an attack-robust solution, as it remains unaffected by attacks on gradient magnitudes. Consequently, several studies \cite{bernstein2018bsignsgd, chen2020distributed, sohn2020election, jin2020stochastic} have focused on enhancing robustness. However, recent developments include new attack methods that can bypass these defense mechanisms \cite{baruch2019little, xie2020fall}, along with the emergence of other backdoor attacks \cite{bagdasaryan2020backdoor, wang2020attack}.

{\noindent \bf Weighted majority voting (WMV):} The WMV method has found widespread application across various domains, such as communication systems \cite{hong2017weighted, kim2019supervised}, crowdsourcing \cite{li2014error, kim2023worker}, and ensemble learning \cite{berend2015finite, kim2023distributed}. Notably, in scenarios involving the transmission of binary information through parallel binary symmetric channels (BSCs), it is established that the WMV decoder, incorporating log-likelihood ratio (LLR) weights, stands as the optimal choice from a maximum likelihood estimation perspective \cite{jeon2018one}. Furthermore, the utility of WMV extends into the realm of federated learning, where it addresses challenges arising from the heterogeneity of data distribution \cite{wu2021fast, li2023revisiting}, as well as mitigating degradation caused by adversarial attacks \cite{jin2020stochastic}. Despite these applications, there remains a notable gap in the literature: specifically, the application of LLR weights on signSGD-based learning algorithms has yet to be explored or studied, to the best of our knowledge.

\section{Preliminaries}
\label{sec:pre}

In this section, we briefly review the classical signSGD-MV algorithm \cite{bernstein2018asignsgd}. We also present an adversarial attack mechanism for the signSGD-MV algorithm, in which adversarial attackers flip the stochastic signs of workers' 
gradients.


\subsection{SignSGD-MV}
\label{sub:signsgd-mv}

 We consider a distributed learning system that consists of one central server and $M$ workers who have their own datasets $\mathcal{D}_m, \forall m \in \mathcal{M}$, where $\mathcal{M}$ is the set of participating workers. These local datasets are the subsets of the global dataset $\mathcal{D}$ with $\mathcal{D} = \bigcup_{m \in \mathcal{M}} \mathcal{D}_m$. Then, the optimization problem of the distributed learning is given by
\begin{align} \label{eqn:DLprob}
    \mathbf{x}^\star = \underset{\mathbf{x} \in \mathbb{R}^N}{\arg \min} \, f (\mathbf{x}) := \underset{\mathbf{x} \in \mathbb{R}^N}{\arg \min} \, \frac{1}{M} \sum_{m \in \mathcal{M}} f_m (\mathbf{x}),
\end{align}
where $f_m (\mathbf{x}) = \mathbb{E}_{\mathbf{d} \sim \mathcal{D}_m} \left[ F \left( \mathbf{x}; \mathbf{d} \right) \right]$ is a local loss function for the worker $m \in \mathcal{M}$. Since the data samples are distributed among several workers, the worker $m$ trains its model to be suitable only for the own dataset $\mathcal{D}_m$. For given model parameter ${\bf x}^t$ at iteration $t$, the worker $m$ computes the stochastic gradient with batch size $B_m$ as
\begin{align}
    \mathbf{g}_m^t := \frac{1}{B_m} \sum_{\mathbf{d} \in \mathcal{B}_m^t} \nabla f_m \left( \mathbf{x}^t; \mathbf{d} \right) \in \mathbb{R}^N,
\end{align}
where $\mathcal{B}_m^t \subset \mathcal{D}_m$. Then, the worker performs one-bit sign quantization for the locally computed stochastic gradient, $\mathsf{sign} \left( \mathbf{g}_m^t \right)$. Subsequently, the sign of the gradient is sent to the server through a band-limited communication network. The server performs the aggregation for $\mathsf{sign} \left( \mathbf{g}_m^t \right)$ using majority voting rule as $\mathsf{sign} \left[ \sum_{m \in \mathcal{M}} \mathsf{sign} \left( \mathbf{g}_m^t \right) \right]$ \cite{bernstein2018asignsgd}.
Then, the server sends the signs of the aggregated gradient to all workers. Lastly, each worker updates the model as follows:
\begin{align} \label{eqn:MV_rule}
    \mathbf{x}^{t+1} = \mathbf{x}^t - \delta \cdot \mathsf{sign} \left[ \sum_{m \in \mathcal{M}} \mathsf{sign} \left( \mathbf{g}_m^t \right) \right],
\end{align}
where $\delta \in \mathbb{R}^+$ is a fixed learning rate parameter. This process is repeated until the model converges, and we call this optimization process as \textit{signSGD-MV}.


\subsection{Adversarial Attacks}
\label{sub:att}

Adversarial attacks in distributed learning gives rise to a substantial threat to the accuracy of training models. This paper specifically focuses on the adversarial attack scenario within the framework of signSGD-MV. Our primary emphasis is on the black-box setting, a scenario wherein attackers are unable to access any datasets from benign workers. This particular setting holds practical significance in numerous distributed learning environments. In this setting, we consider two types of adversarial attacks.

{\bf Sign inversion attack:} Building upon the earlier research conducted by \cite{bernstein2018bsignsgd} and \cite{jin2020stochastic}, a potential adversarial attack can be envisioned, wherein the signs of locally computed gradient are inverted. This particular attack scenario is denoted as the sign-inversion attack (SIA).

{\bf Stochastic sign flip attack:} Inspired by the \textit{Gaussian Byzantine} attack discussed in \cite{blanchard2017machine}, an alternative attack mechanism involves the stochastic flipping of the sign information associated with one-bit gradient information. This attack is hereby referred to as the stochastic sign flip attack (SSFA). Specifically, we characterize the stochastic sign flip attack by introducing a sign-flipping probability parameter, denoted as $r\in [0, 1]$. The set of compromised workers who have been subjected to adversarial attacks is represented by $\mathcal{L} \subset \mathcal{M}$, where $L$ denotes the cardinality of the set. To elaborate, we define the stochastic sign flip attack method as follows: for each compromised worker $\ell \in \mathcal{L}$, the sign information of the $n$th coordinate is stochastically flipped with probability $r$. This can be expressed as follows:
\begin{align} \label{eqn:att_rule}
    \mathsf{sign} \left( \tilde{g}_{\ell, n}^t \right) = 
    \begin{cases}
        \mathsf{sign} \left( g_{\ell, n}^t \right), & \text{w.p. } 1-r \\
        -\mathsf{sign} \left( g_{\ell, n}^t \right), & \text{w.p. } r
    \end{cases},
\end{align} 
where $\tilde{g}_{\ell, n}^t$ is the sign-flipped gradient by the adversarial attacks. Notably, the SSFA boils down to the SIA when $r=1$.


\subsection{Coding-Theoretical Interpretation of signSGD-MV}
\label{sub:interpret}

{\bf Upper bound of signSGD-MV:} Exploring the learning performance of signSGD-MV becomes particularly insightful when we analyze its upper bound. This upper bound is realized when two key conditions are met: i) all workers $m \in \mathcal{M}$ collaboratively utilize the complete local datasets, i.e., $\mathcal{D}_m = \mathcal{D}$, and ii) the gradient computation involves the use of the full-batch size as $B_m = |\mathcal{D}_m|$ for all $m \in \mathcal{M}$. Under these two ideal cases, every worker is empowered to compute the true gradient $\bar{\mathbf{g}}^t$ at each iteration $t$ as
\begin{align}
    \bar{\mathbf{g}}^t = \frac{1}{|\mathcal{D}|} \sum_{\mathbf{d} \in \mathcal{D}} \nabla F \left( \mathbf{x}^t; \mathbf{d} \right).
\end{align}
We denote the true sign of $n$th coordinate as
\begin{align}
    U_n^t = \mathsf{sign} \left( \bar{g}_n^t \right)
\end{align}
for $n\in [N]$. Then, the worker updates the model as
\begin{align}
    x_n^{t+1} = x_n^t - \delta \cdot U_n^t.
\end{align}
The model update technique utilizing the true gradient exhibits a more rapid convergence rate compared to the conventional signSGD-MV, which relies on one-bit stochastic gradient information. From this ideal case, it is important to decode the true sign of the gradient $U_n^t$ in every iteration to speed up the convergence rate. We call this true sign of the gradient $U_n^t$ for all coordinate $n \in [N]$ as \textit{message bits} to be recovered from the aggregation.


{\bf Communication model:} We now interpret the one-bit stochastic gradient computation process as a communication process. Under the limited data knowledge $\mathcal{D}_m \subset \mathcal{D}$ and batch sizes $B_m \le \left| \mathcal{D}_m \right|$, worker $m\in \mathcal{M}$ computes the sign of the stochastic gradient as
\begin{align}
   Y_{m,n}^t={\sf sign}\left( {g}_{m,n}^t \right).
\end{align}
This stochastic sign is different from the true sign of the gradient $U_n^t$. We model this sign mismatch effect with a lens through a communication problem. To be specific, all workers send the true sign of the gradient $U_n^t$ through $M$ parallel \textit{binary symmetric channels} (BSCs). Then, the server receives $M$ independent noisy message bits, i.e., $\mathbf{Y}_n^t = \left[ Y_{1,n}^t, \cdots, Y_{M,n}^t \right]$. In this modeling, the cross-over probability of the BSC, $p_{m,n}^t$ is defined as 
\begin{align}
    p_{m,n}^t = \mathbb{P} \left[ Y_{m,n}^t \ne U_n^t \right]. 
\end{align}
After the server receives all the workers' gradient signs, the server decodes the observations $\mathbf{Y}_n^t$ by using an arbitrary aggregation function $A \left( \mathbf{Y}_n^t \right)$, and we denote the decoded sign as $\hat{U}_n^t$. To mimic the upper bound performance, it is important to design the aggregation function that minimizes the decoding error probability:
\begin{align}
    p_{\mathsf{E}, n}^t = \mathbb{P} \left[ \hat{U}_n^t \ne U_n^t \right].
\end{align}

{\bf Effect of the adversarial attacks in the BSC model:} Utilizing our communication model, we offer insights into how adversarial attacks impact the alteration of cross-over probabilities in BSCs. The SIA simply alters the sign of the locally computed gradient among certain workers. Subsequently, when worker $\ell \in \mathcal{L}$ is subjected to the SIA, we can express the effect as an equivalent adjustment in the cross-over probability. Specifically, the probability $p_{\ell,n}^t$ is transformed to $\tilde{p}_{\ell, n}^t = 1 - p_{\ell,n}^t$ for all $n\in [N]$. The SSFA introduces a probability parameter $r \in [0, 1]$, and modifies the locally computed gradient according to \eqref{eqn:att_rule} by changing its sign with this probability. This operation can be conceptualized as incorporating an additional BSC in a cascade fashion. Specifically, let the flipped sign of the stochastic gradient be denoted as $\tilde{Y}_{\ell, n}^t$ for the coordinate $n \in [N]$. Then, the overall cross-over probability of the two consecutively connected BSCs can be derived as:
\begin{align} \label{eqn:att_prob}
    \tilde{p}_{\ell, n}^t = \mathbb{P} \left[ \tilde{Y}_{\ell, n}^t \ne U_n^t \right] = p_{\ell, n}^t + r \left( 1 - 2 p_{\ell, n}^t \right),
\end{align} 
which can also express that of SIA. Consequently, both SIA and SSFA mechanism can be understood by the changes in the cross-over probabilities of effective BSCs. This interpretation facilitates the establishment of a unified convergence rate applicable to these attack scenarios.


\section{SignSGD with Federated Defense} \label{sec:protect}

In this section, we put forth a novel distributed learning algorithm called signSGD with \textit{federated defense} (signSGD-FD). SignSGD-FD and conventional signSGD-MV share identical algorithm procedure, differing solely in the gradient aggregation method employed at the server. Consequently, we shall focus on explaining the aggregation technique applied to the sign gradient information $\mathbf{Y}_n^t, \forall n \in [N]$.



\subsection{Algorithm}
Under the premise that the cross-over probabilities of all workers, $p_{1,n}^t, \cdots, p_{M,n}^t$, are perfectly known at the server, the optimal aggregation method is to perform the maximum likelihood (ML) decoding. To accomplish ML decoding, the server computes the log-likelihood ratio (LLR) as
\begin{align}
    \ln \frac{\mathbb{P} \left[ \left. \mathbf{Y}_n^t \right| U_n^t = +1 \right]}{\mathbb{P} \left[ \left. \mathbf{Y}_n^t \right| U_n^t = -1 \right]} = \sum_{m \in \mathcal{M}} \ln \frac{1 - p_{m,n}^t}{p_{m,n}^t} Y_{m,n}^t.
\end{align}
As a result, the optimal aggregation boils down to the weighted majority voting (WMV) as
\begin{align} \label{eqn:WMV_rule}
    \hat{U}_{n}^t = \mathsf{sign} \left( \sum_{m \in \mathcal{M}} w_{m,n}^t Y_{m,n}^t \right),
\end{align}
where $w_{m,n}^t = \ln \frac{1 - p_{m,n}^t}{p_{m,n}^t}$ is the $n$th coordinate LLR weight for the worker $m \in \mathcal{M}$. Unfortunately, obtaining the true $p_{m,n}^t$ is an insurmountable task due to the server's inability to access all the data samples from workers. Nevertheless, we can estimate these probabilities from the federated defense mechanism. The key idea behind federated defense lies in leveraging decoding results to estimate the cross-over probabilities $p_{m,n}^t$ over iterations. To elucidate, during the initial phase $t \le T_\mathsf{in}$, the server employs an empirical approach to estimate the probability of computing errors by counting the instances of sign errors across all coordinates $n \in [N]$, and comparing them with the decoding results as
\begin{align} \label{eqn:est_prob_init}
    \hat{p}_{m,n}^{t+1} = \frac{\sum_{i=1}^t \sum_{n=1}^N \mathbf{1}_{\left[ Y_{m,n}^i \ne \hat{U}_{n}^i \right]}}{Nt},  
\end{align}
where $\mathbf{1}_{[\cdot]}$ is an indicator function. Here, the decoding results are obtained by the WMV decoding, but with the estimated LLR weights as
\begin{align} \label{eqn:FD_rule}
    \hat{U}_{\mathsf{FD}, n}^t = \mathsf{sign} \left( \sum_{m \in \mathcal{M}} \hat{w}_{m,n}^t Y_{m,n}^t \right).
\end{align}
From the estimated cross-over probabilities in \eqref{eqn:est_prob_init}, the LLR weight $\hat{w}_{m,n}^{t+1}$ is updated as
\begin{align} \label{eqn:est_LLR}
    \hat{w}_{m,n}^{t+1} = \ln \frac{1 - \hat{p}_{m,n}^{t+1}}{\hat{p}_{m,n}^{t+1}}.
\end{align}
After the initial phase $t > T_\mathsf{in}$, the cross-over probability estimation rule is changed to recursively update $\hat{p}_{m,n}^t$ in parallel for each coordinate as
\begin{align}
    \hat{p}_{m,n}^{t+1} = \frac{T_\mathsf{in}}{t} \hat{p}_{m,n}^{T_\mathsf{in}} + \frac{t - T_\mathsf{in}}{t} \frac{\sum_{i=T_\mathsf{in} + 1}^t \mathbf{1}_{\left[ Y_{m,n}^i \ne \hat{U}_{n}^i \right]}}{t - T_\mathsf{in}}, \label{eq:prob_eff}
\end{align} 
where $\hat{p}_{m,n}^{T_\mathsf{in}}$ is computed by following \eqref{eqn:est_prob_init}. Consequently, the model is updated as
\begin{align}
    \mathbf{x}^{t+1} = \mathbf{x}^t - \delta \cdot \hat{U}_{\mathsf{FD}, n}^t.
\end{align}
The entire algorithm is summarized in Algorithm \ref{alg:signSGD-FD}.


\subsection{Remarks}

{\bf Universality for adversarial attacks:} The proposed signSGD-FD exhibits versatility in addressing a broad spectrum of adversarial attack scenarios. Its adaptability is particularly evident in the nuanced modeling of cross-over probabilities based on distinct adversarial attack mechanisms. In the context of the SIA scenario, the cross-over probability undergoes a transformation from $p_{\ell,n}^t$ to ${\tilde p}_{\ell,n}^t=1-p_{\ell,n}^t$ when worker $\ell \in \mathcal{L}$ is targeted. Meanwhile, in the SSFA scenario, which can generalize the SIA method, the cross-over probability is precisely modeled as the expression given by \eqref{eqn:att_prob}. A noteworthy aspect of the proposed FD is its independence from any prior knowledge regarding adversarial attack scenarios. Notably, it eliminates the need to estimate the sign-flipping probability $r$ under the SSFA. The only requisite information is the estimation of effective cross-over probabilities $\hat{p}_{m,n}^t$ in \eqref{eq:prob_eff}. Consequently, our signSGD-FD emerges as a versatile solution, applicable across diverse adversarial attack scenarios.

{\bf Harnessing compromised workers:} Our signSGD-FD algorithm leverages the gradients of compromised workers for aggregation. Specifically, following the cross-over probability estimation process in \eqref{eqn:est_prob_init} and \eqref{eq:prob_eff}, the server can pinpoint the compromised workers by identifying workers whose cross-over probabilities exceed 1/2. Typically, during gradient aggregation, the server can eliminate the local gradients from these identified adversarial workers. However, our federated defense mechanism demonstrates that this elimination strategy is notably sub-optimal. To achieve optimal ML decoding performance, it is crucial to utilize the cross-over probabilities of all workers. These probabilities are imperative because the sign of the estimated LLR weights of compromised workers can automatically change if ${\hat p}_{m,n}^t >1/2$. This counter-intuitive result will be verified from the convergence analysis in the subsequent section.

\begin{algorithm}[tb]
   \caption{signSGD-FD}
   \label{alg:signSGD-FD}
\begin{algorithmic}
   \STATE {\bfseries Input:} Initial model $\mathbf{x}^1$, the number of workers $M$, worker $m$'s batch size $B_m$, learning rate $\delta$, initial weight $\hat{w}_{m,n}^1 = 1$, initial phase duration $T_\mathsf{in}$, total iteration $T$
   \vspace{0.2em} 
   \FOR{$t = 1: T$}
   \STATE \vspace{-0.5em} \hrulefill
   \FOR{{\bf each worker} $m \in \mathcal{M}$}
   \STATE {\bf Compute} $\mathbf{g}_m^t$ with batch size $B_m$
   \STATE {\bf Encode} $Y_{m,n}^t = \mathsf{sign} \left( g_{m,n}^t \right), \forall n \in [N]$
   \STATE {\bf Send} $\mathsf{sign} \left( \mathbf{g}_m^t \right) = \left[ Y_{m, 1}^t, \cdots, Y_{m, N}^t \right]$ to {\bf server}
   \ENDFOR
   \\ \vspace{-0.5em} \hrulefill

   \FOR{{\bf each worker} $\ell \in \mathcal{L}$ {\bf attackers}}
   \vspace{0.1em}
   \STATE {\bf Manipulate} $\tilde{Y}_{\ell, n}^t \! = \! 
   \begin{cases}
        Y_{\ell, n}^t, \!\!\!\!  & \text{w.p. } 1 \!-\! r \\
        -Y_{\ell, n}^t, \!\!\!\!  & \text{w.p. } r
   \end{cases}\!, \forall n \in [N]$ 
   \vspace{0.1em}
   \STATE {\bf Send} $\mathsf{sign} \left( \tilde{\mathbf{g}}_\ell^t \right) = \left[ \tilde{Y}_{\ell, 1}^t, \cdots, \tilde{Y}_{\ell, N}^t \right]$ to {\bf server}
   \ENDFOR
   \\ \vspace{-0.5em} \hrulefill

   \FOR{{\bf each coordinate} $n = 1: N$ {\bf server}}
   \STATE {\bf Decode} \\ \vspace{0.1em}
   $\hat{U}_n^t = \mathsf{sign} \left( \sum_{m \in \mathcal{M} \backslash \mathcal{L}} \hat{w}_{m,n}^t Y_{m,n}^t \!+\! \sum_{\ell \in \mathcal{L}} \hat{w}_{\ell, n}^t \tilde{Y}_{\ell, n}^t \right)$
   \FOR{$m \in \mathcal{M}$}
   \STATE {\bf Estimate} \\
   $ \hat{p}_{m,n}^t = 
   \begin{cases}
       \frac{\sum_{i=1}^t \sum_{n=1}^N \mathbf{1}_{\left[ Y_{m,n}^i \ne \hat{U}_n^i \right]}}{Nt}, & \text{if } t \le T_\mathsf{in} \\
       \frac{\sum_{i=1}^t \mathbf{1}_{\left[ Y_{m,n}^i \ne \hat{U}_n^i \right]}}{t}, & \text{o.w.}
   \end{cases}$
   \STATE {\bf Update} $\hat{w}_{m,n}^t = \ln \frac{1 - \hat{p}_{m,n}^t}{\hat{p}_{m,n}^t}$
   \ENDFOR
   \STATE {\bf Send} $\hat{U}_n^t$ to {\bf all workers} $m \in \mathcal{M}$
   \ENDFOR
   \\ \vspace{-0.5em} \hrulefill

   \FOR{{\bf each worker} $m \in \mathcal{M}$}
   \STATE {\bf Update} $x_n^{t+1} = x_n^t - \delta \cdot \hat{U}_n^t, \,\forall n \in [N]$
   \ENDFOR
   \\ \vspace{-0.5em} \hrulefill
   \ENDFOR
\end{algorithmic}
\end{algorithm}

\section{Convergence Analysis}
\label{sec:convergence}

In this section, we provide the convergence analysis for signSGD-FD in the presence of stochastic sign flip attacks. 


\subsection{Assumptions} \label{sub:ass}

Before analyzing the convergence guarantee, we first present some assumptions used in the analysis:

\begin{assumption}[Lower bound] \label{ass:1}
    For all $\mathbf{x} \in \mathbb{R}^N$ and some local minimum points $\mathbf{x}^\star$, we have an objective value as
    \begin{align}
        f(\mathbf{x}) \ge f \left( \mathbf{x}^\star \right) = f^\star.
    \end{align}
\end{assumption}

\begin{assumption}[Coordinate-wise smoothness] \label{ass:2}
    For all $\mathbf{x}, \mathbf{y} \in \mathbb{R}^N$, there exists a vector with non-negative constants $\mathbf{L} = \left[ L_1, \ldots, L_N \right]$ that satisfies
    \begin{align}
        \left| f(\mathbf{y}) \! - \! f(\mathbf{x}) \! - \! \left\langle \nabla f(\mathbf{x}), \mathbf{y} \! - \! \mathbf{x} \right\rangle \right| \! \le \! \sum_{n=1}^N \! \frac{L_n}{2} \! \left( y_n \! - \! x_n \right)^2 \!\! .
    \end{align}
\end{assumption}



Assumption \ref{ass:1} is required for the convergence to local minima, and Assumption \ref{ass:2} indicates the Lipschitz condition of the objective function. Assumptions \ref{ass:1} and \ref{ass:2} are commonly used for the convergence analysis of learning algorithms as in \cite{li2019convergence, bernstein2018asignsgd}, but with a coordinate-wise fashion.


\subsection{Convergence Analysis without Attacks} \label{sub:anlys_noattack}

Under these mild assumptions, the convergence rate of signSGD using an arbitrary binary aggregation can be derived as the theorem below.

\begin{theorem}[Universal convergence rate] \label{thm:1}
    Let $\hat{U}_n^t = A \left( \mathbf{Y}_n^t \right) \in \{-1,+1\}$ be a decoded gradient sign for $n$th coordinate at iteration $t$. We define the maximum of sign decoding error probability over all coordinates and iterations as
    \begin{align}
        P_\mathsf{E}^\mathsf{max} = \underset{n \in [N], t \in [T]}{\max} \mathbb{P} \left[ \hat{U}_n^t \ne U_n^t \right].
    \end{align}
    With a fixed learning parameter $\delta = \sqrt{\frac{2\left( f^1 - f^\star \right)}{T \lVert \mathbf{L} \rVert_1}}$, the convergence rate of signSGD-type algorithms is given by
    \begin{align}
        \mathbb{E} \left[ \frac{1}{T} \! \sum_{t=1}^{T} \lVert \bar{\mathbf{g}}^t \rVert_1 \right] \le \frac{1}{1 \!-\! 2 P_\mathsf{E}^\mathsf{max}} \sqrt{\frac{2 \left( f^1 \!-\! f^\star \right) \lVert \mathbf{L} \rVert_1}{T}},
    \end{align}
    for $P_\mathsf{E}^\mathsf{max} < \frac{1}{2}$.
\end{theorem}
The convergence rate in Theorem \ref{thm:1} holds for an arbitrary sign decoding function $A \left( \mathbf{Y}_n^t \right)$. From Theorem \ref{thm:1}, we observe that the convergence rates of signSGD-style algorithms, including signGD, have an order $\mathcal{O} \left( \frac{1}{\sqrt{T}} \right)$. More importantly, the convergence rate of signSGD-based algorithms improves as decreasing the maximum decoding error probability. From this observation, our focus shifts to establishing an upper bound for the decoding error probability when applying the proposed signSGD-FD algorithm.


\begin{theorem}[Decoding error bound of signSGD-FD] \label{thm:2}
    For every $n \in [N]$, $m \in \mathcal{M}$, and $t \in [T]$, suppose the ratio between the estimated and true LLR weights are bounded with some constants  $\delta_\mathsf{max} \in \mathbb{R}^+$ and $\delta_\mathsf{min} \in \mathbb{R}^+$ as
    \begin{align} \label{eqn:weight_delta}
        1 - \delta_\mathsf{min} \le \frac{\hat{w}_{m,n}^t}{w_{m,n}^t} \le 1 + \delta_\mathsf{max}.
    \end{align}
Then, the gradient sign decoding error probability when applying the FD aggregation in \eqref{eqn:FD_rule} is upper bounded by
    \begin{align} \label{eqn:WMV_errorbound}
        P_\mathsf{E}^\mathsf{FD} \le \exp \left[ -M \left( \frac{1 - \delta_\mathsf{min}}{1 + \delta_\mathsf{max}} \right) \gamma_\mathcal{M}^\mathsf{WMV} \right],
    \end{align}
    where $\gamma_\mathcal{M}^\mathsf{WMV} = \frac{1}{M} \sum_{m \in \mathcal{M}} \frac{1}{2} \left( \frac{1}{2} - p_{m,n}^t \right) \ln \frac{1 - p_{m,n}^t}{p_{m,n}^t}$ is the error exponent of the perfect WMV aggregation for the entire worker set $\mathcal{M}$.
\end{theorem}
From Theorem \ref{thm:2}, we observe that the decoding error bound exponentially decreases with the number of workers $M$. The error exponent $\gamma_\mathcal{M}^\mathsf{WMV}$ determines how quickly the error probability diminishes as increasing $M$. In addition, the performance loss of FD aggregation due to weight uncertainty results in an error exponent reduction of $\frac{1 - \delta_\mathsf{min}}{1 + \delta_\mathsf{max}} \le 1$. This observation confirms that the accurate $\hat{p}_{m,n}^t$ estimation helps to reduce the decoding error probability.


\begin{theorem}[Decoding error bound of signSGD-MV] \label{thm:3}
    Suppose the server performs the MV aggregation in \eqref{eqn:MV_rule}, i.e., $\hat{U}_{\mathsf{MV}, n}^t = \mathsf{sign} \left( \sum_{m \in \mathcal{M}} Y_{m,n}^t \right)$. Then, the decoding error probability is upper bounded by
    \begin{align}
        P_\mathsf{E}^\mathsf{MV} \le \exp \left( -M \gamma_\mathcal{M}^\mathsf{MV} \right),
    \end{align}
    where $\gamma_\mathcal{M}^\mathsf{MV} = \bar{p}_n^t - \frac{1}{2} \ln \left( 2e \bar{p}_n^t \right)$ is the error exponent of the MV aggregation, and $\bar{p}_n^t = \frac{1}{M} \sum_{m \in \mathcal{M}} p_{m,n}^t$ is the average of workers' cross-over probabilities.
\end{theorem}

The error bound of MV decoder established in Theorem \ref{thm:3} follows an exponentially decreasing trend with $M$, same with the FD aggregation. The error exponent $\gamma_\mathcal{M}^\mathsf{MV}$ is expressed as the average of workers' computing error probabilities, and this term determines the decoding performance.

\subsection{Upper Bounds of Decoding Errors under Attacks} \label{sub:anlys_attack}
We establish the upper bounds of the decoding error probability of signSGD-FD under the SSFA. When the SSFA is considered, as explained in \eqref{eqn:att_prob}, the effective cross-over probability ${\tilde p}_{\ell,n}^t$ increases, which leads to an increase of decoding error probability. The theorem below elucidates the deterioration of the FD aggregation in the presence of SSFAs.

\begin{theorem}[Decoding error bound for signSGD-FD under the SSFA]\label{thm:4}
    Suppose the server performs the FD aggregation, and the workers in $\mathcal{L} \subset \mathcal{M}$ with cardinality $L$ are contaminated by the stochastic sign flip attacks with probability $r$. Then, the decoding error probability is upper bounded by
    \begin{align} \label{eqn:WMVbound_att}
        P_\mathsf{E}^\mathsf{FD} \le \exp \left[ -(M - L) \left( \frac{1 - \delta_\mathsf{min}}{1 + \delta_\mathsf{max}} \right) \tilde{\gamma}_{\mathcal{M}, \mathcal{L}}^\mathsf{WMV} \right], 
    \end{align}
    where $\tilde{\gamma}_{\mathcal{M}, \mathcal{L}}^\mathsf{WMV} \! = \! \frac{1}{M-L} \! \left[ M \gamma_\mathcal{M}^\mathsf{WMV} \!\! - \! L \gamma_\mathcal{L}^\mathsf{WMV} \!\! + \! \sum_{\ell \in \mathcal{L}} g \! \left( \tilde{p}_{\ell, n}^t \right) \right]$ is the modified error exponent of the FD aggregation with a specific function $g(p) = \frac{1}{2} \left( \frac{1}{2} - p \right) \ln \frac{1-p}{p}$.
\end{theorem}
Theorem \ref{thm:4} shows how the decoding error bound of signSGD-FD changes by the SSFA with probability $r$. First, the effective number of workers decreases from $M$ to $M-L$, which can increase the decoding error bound. However, $\tilde{\gamma}_{\mathcal{M}, \mathcal{L}}^\mathsf{WMV} $ is greater than ${\gamma}_{\mathcal{M}, \mathcal{L}}^\mathsf{WMV} $, which leads to the decrease of the decoding error bound. To provide a more clear intuition on the two opposite effects, we provide the following corollary for some special cases of the sign flip probability.


\begin{corollary}[Special case] \label{cor:1}
    Under the SSFA with the sign flip probability $r\in \{1, \frac{1}{2}\}$, the decoding error probablity is upper bounded by
    \begin{align}
        P_\mathsf{E}^\mathsf{WMV} \! \leq \! 
        \begin{cases}
            \exp \left[ -M \left( \frac{1 - \delta_\mathsf{min}}{1 + \delta_\mathsf{max}} \right)  \gamma_\mathcal{M}^\mathsf{WMV} \right], & \text{if } r = 1 \\
            \exp \left[ - \left( M-L \right)\left( \frac{1 - \delta_\mathsf{min}}{1 + \delta_\mathsf{max}} \right)   \gamma_\mathcal{M}^\mathsf{WMV} \right], & \text{if } r = \frac{1}{2}
        \end{cases}.
    \end{align} 
\end{corollary}

Corollary \ref{cor:1} clearly shows that the decoding error probability bound does not change under the SSFA with the sign flip probability $r=1$, i.e., the SIA. This result is counter-intuitive because the adversarial attacks have shown the degradation of the convergence rate. However, our result confirms that the convergence rate remains unchanged if we sagaciously harness the gradient information sent by the adversarial workers with proper weights. It is also noted that when $r=\frac{1}{2}$, the exponent term decreases from $M$ to $M-L$, which slows down the convergence rate. From this result, we also observe that the worst-case attack scenario is to use the sign flip probability of $r=\frac{1}{2}$. This observation aligns with our intuition that the uniformly random gradient sign flipping does not allow the server to exploit the gradient sign information for decoding.

To better appreciate the distinctions, it is informative to juxtapose the decoding error bound of signSGD-MV within the framework of SSFA with $r$ against that of signSGD-FD.

\begin{theorem}[Decoding error bound of signSGD-MV] \label{thm:5}
    Suppose $L$ workers are under the SSFA with $r$. Then, the decoding error probability of the signSGD-MV algorithm is upper bounded by
    \begin{align}
        P_\mathsf{E}^\mathsf{MV} \le \exp \left[ - \left( M-2rL \right) \tilde{\gamma}_{\mathcal{M}, \mathcal{L}}^\mathsf{MV} \right],
    \end{align}
    where $\tilde{\gamma}_{\mathcal{M}, \mathcal{L}}^\mathsf{MV} = \frac{1}{M-2rL} \left[ M \gamma_\mathcal{M}^\mathsf{MV} - 2rL \gamma_\mathcal{L}^\mathsf{MV} + \epsilon_\mathcal{L} (r) \right]$ is the modified error exponent of MV aggregation expressed with a sufficiently small parameter $\epsilon_\mathcal{L} (r)$.
\end{theorem}

Theorem \ref{thm:5} shows that the decoding error probability worsens as the sign flip probability $r$ increases. Unlike our signSGD-FD method, when $r=1$, the convergence rate of signSGD-MV is significantly degraded by SSFA because the decoding error bound increases considerably. Nevertheless, signSGD-MV can achieve the identical convergence rate with signSGD-FD under the SSFA when $r=\frac{1}{2}$. As a result, signSGD-FD provides a theoretical guarantee of superior convergence rate than signSGD-MV under the SSFA when $r>\frac{1}{2}$. We also empirically observe from numerical experiments that signSGD-FD provides a faster convergence rate than signSGD-MV under the SSFA, even when $r=\frac{1}{2}$.  



\section{Experiments} \label{sec:experiments}

This section presents the experimental results on the image classification to verify the learning performance of signSGD-FD compared to other distributed learning algorithms.

\subsection{Settings} \label{sub:set}

{\bf Datasets \& Training models.} The real datasets used for image classification simulation are MNIST \cite{lecun1998gradient}, CIFAR-10, and CIFAR-100 \cite{krizhevsky2009learning} datasets. For the learning models, we adopt a convolutional neural network (CNN) \cite{lecun1998gradient} model for the MNIST dataset, and a ResNet-56 \cite{he2016deep} model for the CIFAR-10 and CIFAR-100 datasets. The number of workers $M$ is fixed to 15, and more details about simulation settings are deferred to Appendix \ref{app:details} and \ref{sub:init_dec}.

{\bf Adversarial attacks.} To evaluate signSGD-FD in the presence of attackers, we introduce the SIA and SSFA method in Section \ref{sub:att}. The sign-flipping probability $r$ is set between 0 and 1, especially focusing on the $r=1$ which is the SIA case and the $r = \frac{1}{2}$ case. We fix the set of workers $\mathcal{L}$ compromised by the attacks during the entire learning.


\subsection{Effects of SIA $(r=1)$} \label{sub:r=1}

\begin{figure}[!t]
\vskip 0.2in
\begin{center}
\subfigure[CIFAR-10 with $r=1$]{%
    \includegraphics[width=0.49\columnwidth]{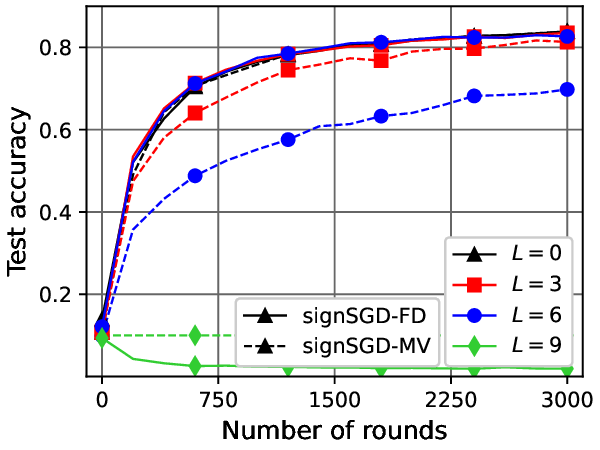}}
\hfill
\subfigure[CIFAR-100 with $r=1$]{%
    \includegraphics[width=0.49\columnwidth]{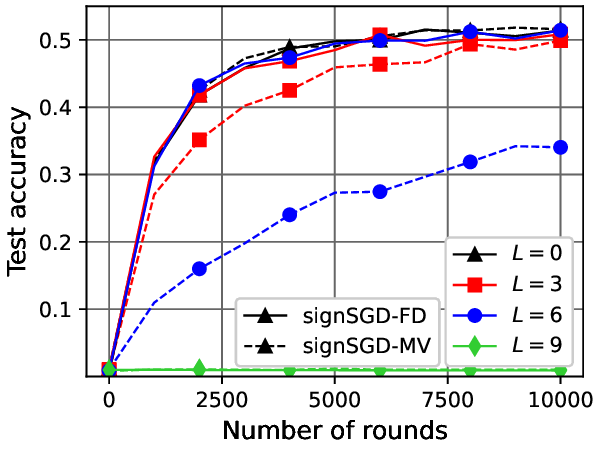}}
\subfigure[CIFAR-10 with $r=\frac{1}{2}$]{%
    \includegraphics[width=0.49\columnwidth]{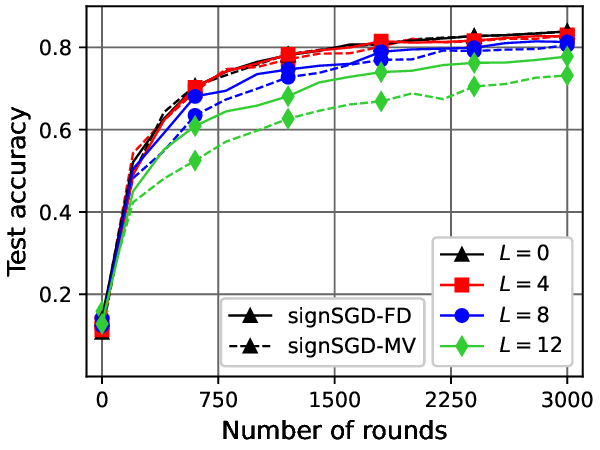}}
\hfill
\subfigure[CIFAR-100 with $r=\frac{1}{2}$]{%
    \includegraphics[width=0.49\columnwidth]{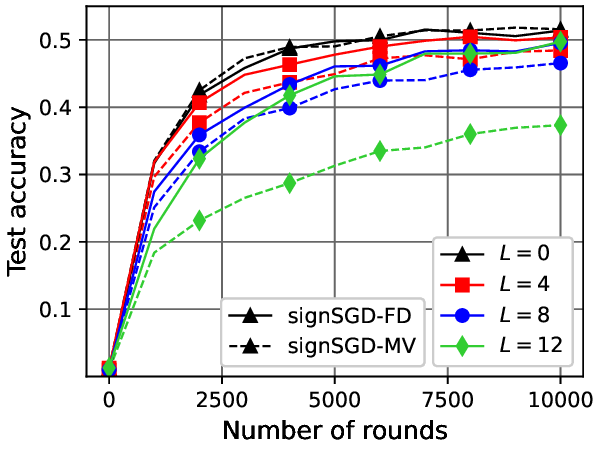}}
\caption{Test accuracy vs. training rounds varying the number of compromised workers $L$.}
\label{fig:1} 
\end{center}
\vskip -0.3in
\end{figure}

In Figure \ref{fig:1}-(a) and (b), we provide the test accuracy assessment of signSGD-FD on the effect of SIA. The general trend of results is that signSGD-MV deteriorates significantly as the number of compromised workers $L$ increases, while signSGD-FD can achieve almost the same accuracy as in the absence of attacks if $L < \frac{M}{2}$. These results convince us that the LLR weights used in the FD aggregation can provide excellent protection of the training models against the SIA method. Meanwhile, it can be seen that signSGD-FD fails to converge in the $L=9$ case. This can be considered that the decoding error probability becomes greater than $\frac{1}{2}$ in this case, making it no longer possible to perform the accurate $\hat{p}_{m,n}^t$ estimation. These observations are consistent with the analyses in Section \ref{sec:convergence}.

\subsection{Effects of SSFA $\left(r = \frac{1}{2}\right)$} \label{sub:r=0.5}


We also evaluate signSGD-FD in the presence of SSFA with $r=\frac{1}{2}$ in Figure \ref{fig:1}-(c) and (d). The test accuracy results show that signSGD-FD still be more robust to the sign flip attacks with $r=\frac{1}{2}$ than signSGD-MV. Nevertheless, signSGD-FD becomes to degrade gradually as the number of compromised workers $L$ increases, which aligns with our analyses in Corollary \ref{cor:1}. A notable point is that both algorithms can learn the models in $L = 12$ case, unlike the SIA case, since the effective number of workers for both algorithms becomes $M-L > 0$.

\subsection{Degradation Trend According to $r$} \label{sub:deg_r}

\begin{figure}[!t]
\vskip 0.2in
\begin{center}
\subfigure[CIFAR-10 dataset]{%
    \includegraphics[width=0.49\columnwidth]{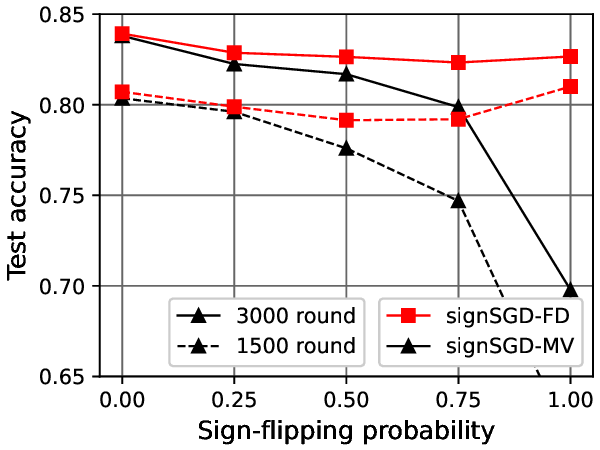}}
\hfill
\subfigure[CIFAR-100 dataset]{%
    \includegraphics[width=0.49\columnwidth]{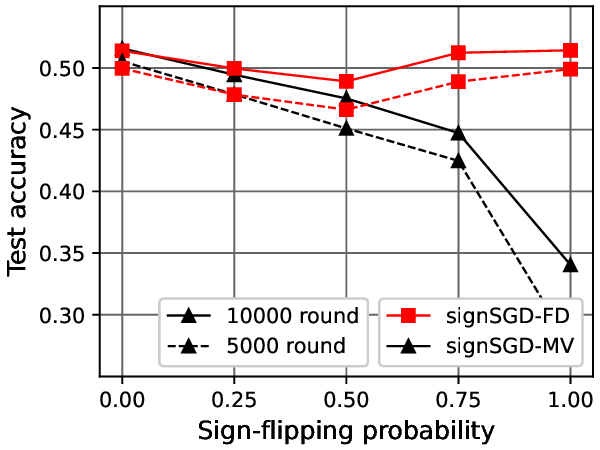}}
\caption{Test accuracy vs. training rounds varying sign-flipping probability $r$.}
\label{fig:2} 
\end{center}
\vskip -0.3in
\end{figure}

The degradation trend according to the sign-flipping probability $r$ is illustrated in Figure \ref{fig:2} with $L=6$. The results for both datasets have the same trend that the test accuracy loss becomes severer as $r$ increases. On the other hand, signSGD-FD achieves the best accuracy when $r=0$ or 1, and the worst accuracy for the $r=\frac{1}{2}$ attack, but not that serious. These results have already predicted through the channel capacity analysis on SSFA.

\subsection{Algorithms Comparison} \label{sub:comp_alg}

\begin{figure}[!t]
\vskip 0.2in
\begin{center}
\subfigure[signSGD-based algorithms]{%
    \includegraphics[width=0.49\columnwidth]{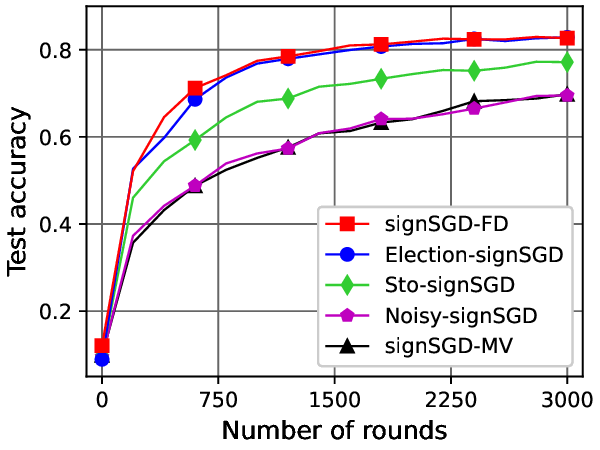}}
\hfill
\subfigure[Commun. costs comparison]{%
    \includegraphics[width=0.49\columnwidth]{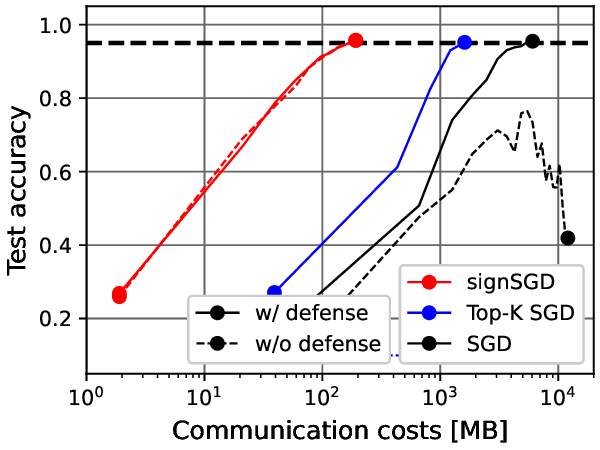}}
\caption{Test accuracy \& communication costs comparison among the attack-robust distributed learning algorithms.}
\label{fig:4} 
\end{center}
\vskip -0.2in
\end{figure}

We finally compare the robustness against attacks by comparing the test accuracy of signSGD-style algorithms, depicted in Figure \ref{fig:4}-(a). Here, we set $L=6$ with the SIA method for this comparison. The details about the compared algorithms are presented in Appendix \ref{app:details}. The comparison results verify that the proposed signSGD-FD algorithm can achieve the highest test accuracy compared to other signSGD-style robust optimizers. Moreover, other algorithms require susceptible parameter settings to obtain high accuracy, but signSGD-FD is not so affected by these settings, which can be observed in Appendix \ref{sub:init_dec}.


Figure \ref{fig:4}-(b) shows each algorithm's communication costs on the MNIST dataset. In this comparison, we adopt \textit{Gaussian Byzantine} attack, and also apply \textit{Multi-Krum} algorithm on SGD and Top-$K$ SGD optimizers to be robust against attacks. Appendix \ref{app:details} illustrates the details of these algorithms. The results demonstrate that signSGD-type algorithms can significantly reduce communication costs by 30x compared to the Multi-Krum algorithm. We can even observe the robustness of signSGD, showing that signSGD-MV can achieve the same accuracy as signSGD-FD. Top-K SGD-based multi-Krum requires 3x fewer costs than SGD but still needs 10x more costs than our proposed algorithm, which reveals the superiority of signSGD-FD.

\section{Conclusion}
\label{sec:conclusion}

In this paper, we have shown that the convergence rate of the signSGD-type algorithm is invariant regardless of the number of adversarial workers, as long as their counts remain lower than that of legitimate workers. This counter-intuitive result is achieved through a new optimizer, signSGD-FD, incorporating the concept of federated defense. Federated defense utilizes learnable weights for weighted majority voting during aggregation. The server dynamically learns these weights based on the workers' reliability estimation of transmitted local gradient information. The weights are then applied to decode the sign of aggregated local gradients, minimizing sign decoding errors. We have also provided a unified convergence rate analysis framework applicable to various adversarial attack scenarios. Experimental results demonstrate that signSGD-FD outperforms traditional signSGD-MV, showcasing a faster convergence rate, especially in the presence of adversarial attacks.


\section*{Impact Statements}

This paper presents work that aims to advance the field of distributed learning under adversarial attacks. Present-day deep neural networks (DNNs) exhibit susceptibility to adversarial attacks, wherein maliciously crafted perturbations applied to input data or model weights can alter or manipulate the classification outcomes. Previously, adversarial attacks have been shown to decrease distributed learning performance. In this paper, however, we show that the distributed learning performance can remain unchanged under some adversarial attacks if we harness the information from adversarial workers in the learning process. This new principle of exploiting adversarial workers can impact the development of many other distributed learning algorithms resilient to adversarial attacks.




\nocite{park2023s}

\bibliography{ICML_arXiv}

\begin{thebibliography}{55}
\providecommand{\natexlab}[1]{#1}
\providecommand{\url}[1]{\texttt{#1}}
\expandafter\ifx\csname urlstyle\endcsname\relax
  \providecommand{\doi}[1]{doi: #1}\else
  \providecommand{\doi}{doi: \begingroup \urlstyle{rm}\Url}\fi

\bibitem[Aji \& Heafield(2017)Aji and Heafield]{aji2017sparse}
Aji, A.~F. and Heafield, K.
\newblock Sparse communication for distributed gradient descent.
\newblock \emph{arXiv preprint arXiv:1704.05021}, 2017.

\bibitem[Alistarh et~al.(2017)Alistarh, Grubic, Li, Tomioka, and
  Vojnovic]{alistarh2017qsgd}
Alistarh, D., Grubic, D., Li, J., Tomioka, R., and Vojnovic, M.
\newblock {QSGD}: Communication-efficient {SGD} via gradient quantization and
  encoding.
\newblock In \emph{Advances in Neural Information Processsing Systems
  (NeurIPS)}, volume~30, pp.\  1709--1720, Long Beach, CA, 2017.

\bibitem[Alistarh et~al.(2018)Alistarh, Allen-Zhu, and
  Li]{alistarh2018byzantine}
Alistarh, D., Allen-Zhu, Z., and Li, J.
\newblock Byzantine stochastic gradient descent.
\newblock In \emph{Advances in Neural Information Processsing Systems
  (NeurIPS)}, volume~31, Montréal, Canada, 2018.

\bibitem[Bagdasaryan et~al.(2020)Bagdasaryan, Veit, Hua, Estrin, and
  Shmatikov]{bagdasaryan2020backdoor}
Bagdasaryan, E., Veit, A., Hua, Y., Estrin, D., and Shmatikov, V.
\newblock How to backdoor federated learning.
\newblock In \emph{Proceedings of the 23rd International Conference on
  Artificial Intelligenec and Statistics (AISTATS)}, volume 108, pp.\
  2938--2948, Online, 2020.

\bibitem[Baruch et~al.(2019)Baruch, Baruch, and Goldberg]{baruch2019little}
Baruch, G., Baruch, M., and Goldberg, Y.
\newblock A little is enough: Circumventing defenses for distributed learning.
\newblock In \emph{Advances in Neural Information Processing Systems
  (NeurIPS)}, volume~32, Vancouver, Canada, 2019.

\bibitem[Basu et~al.(2019)Basu, Data, Karakus, and Diggavi]{basu2019qsparse}
Basu, D., Data, D., Karakus, C., and Diggavi, S.
\newblock Qsparse-local-{SGD}: Distributed {SGD} with quantization,
  sparsification and local computations.
\newblock In \emph{Advances in Neural Information Processsing Systems
  (NeurIPS)}, volume~32, pp.\  14695--14706, Vancouver, Canada, 2019.

\bibitem[Berend \& Kontorovich(2015)Berend and Kontorovich]{berend2015finite}
Berend, D. and Kontorovich, A.
\newblock A finite sample analysis of the naive bayes classifier.
\newblock \emph{Journal of Machine Learning Research}, 16\penalty0
  (1):\penalty0 1519--1545, 2015.

\bibitem[Bernstein et~al.(2018)Bernstein, Wang, Azizzadenesheli, and
  Anandkumar]{bernstein2018asignsgd}
Bernstein, J., Wang, Y.-X., Azizzadenesheli, K., and Anandkumar, A.
\newblock sign{SGD}: Compressed optimisation for non-convex problems.
\newblock In \emph{Proceedings of the 35th International Conference on Machine
  Learning (ICML)}, pp.\  560--569, Stockholm, Sweden, 2018.

\bibitem[Bernstein et~al.(2019)Bernstein, Zhao, Azizzadenesheli, and
  Anandkumar]{bernstein2018bsignsgd}
Bernstein, J., Zhao, J., Azizzadenesheli, K., and Anandkumar, A.
\newblock sign{SGD} with majority vote is communication efficient and fault
  tolerant.
\newblock In \emph{Proceedings of the 7th International Conference on Learning
  Representations}, New Orleans, LA, 2019.

\bibitem[Blanchard et~al.(2017)Blanchard, El~Mhamdi, Guerraoui, and
  Stainer]{blanchard2017machine}
Blanchard, P., El~Mhamdi, E.~M., Guerraoui, R., and Stainer, J.
\newblock Machine learning with adversaries: Byzantine tolerant gradient
  descent.
\newblock In \emph{Advances in Neural Information Processsing Systems},
  volume~30, Long Beach, CA, 2017.

\bibitem[Bottou(2010)]{bottou2010large}
Bottou, L.
\newblock Large-scale machine learning with stochastic gradient descent.
\newblock In \emph{Proceedings of the 19th International Conference on
  Computational Statistics (COMPSTAT)}, pp.\  177--186, Paris, France, 2010.

\bibitem[Chen et~al.(2020)Chen, Chen, Sun, Wu, and Hong]{chen2020distributed}
Chen, X., Chen, T., Sun, H., Wu, S.~Z., and Hong, M.
\newblock Distributed training with heterogeneous data: Bridging median-and
  mean-based algorithms.
\newblock In \emph{Advances in Neural Information Processsing Systems
  (NeurIPS)}, volume~33, pp.\  21616--21626, Online, 2020.

\bibitem[Dean et~al.(2012)Dean, Corrado, Monga, Chen, Devin, Mao, Ranzato,
  Senior, Tucker, Yang, Le, and Ng]{dean2012large}
Dean, J., Corrado, G., Monga, R., Chen, K., Devin, M., Mao, M., Ranzato, M.,
  Senior, A., Tucker, P., Yang, K., Le, Q., and Ng, A.
\newblock Large scale distributed deep networks.
\newblock In \emph{Advances in Neural Information Processing Systems (NIPS)},
  volume~25, pp.\  1223--1231, Lake Tahoe, NV, 2012.

\bibitem[Gandikota et~al.(2021)Gandikota, Kane, Maity, and
  Mazumdar]{gandikota2021vqsgd}
Gandikota, V., Kane, D., Maity, R.~K., and Mazumdar, A.
\newblock vq{SGD}: Vector quantized stochastic gradient descent.
\newblock In \emph{Proceedings of the 24th International Conference on
  Artificial Intelligence and Statistics (AISTATS)}, pp.\  2197--2205, Online,
  2021.

\bibitem[Guerraoui et~al.(2018)Guerraoui, Rouault, et~al.]{guerraoui2018hidden}
Guerraoui, R., Rouault, S., et~al.
\newblock The hidden vulnerability of distributed learning in {B}yzantium.
\newblock In \emph{Proceedings of the 35th International Conference on Machine
  Learning (ICML)}, pp.\  3521--3530, Stockholm, Sweden, 2018.

\bibitem[He et~al.(2016)He, Zhang, Ren, and Sun]{he2016deep}
He, K., Zhang, X., Ren, S., and Sun, J.
\newblock Deep residual learning for image recognition.
\newblock In \emph{Proceedings of the IEEE Conference on Computer Vision and
  Pattern Recognition (CVPR)}, pp.\  770--778, Las Vegas, NV, 2016.

\bibitem[Hong et~al.(2017)Hong, Kim, and Lee]{hong2017weighted}
Hong, S.-N., Kim, S., and Lee, N.
\newblock A weighted minimum distance decoding for uplink multiuser {MIMO}
  systems with low-resolution {ADC}s.
\newblock \emph{IEEE Transactions on Communications}, 66\penalty0 (5):\penalty0
  1912--1924, 2017.

\bibitem[H{\"o}nig et~al.(2022)H{\"o}nig, Zhao, and
  Mullins]{honig2022dadaquant}
H{\"o}nig, R., Zhao, Y., and Mullins, R.
\newblock {DAdaQuant}: Doubly-adaptive quantization for communication-efficient
  federated learning.
\newblock In \emph{Proceedings of the 39th International Conference on Machine
  Learning (ICML)}, pp.\  8852--8866, Baltimore, MD, 2022.

\bibitem[Jeon et~al.(2018)Jeon, Lee, Hong, and Heath]{jeon2018one}
Jeon, Y.-S., Lee, N., Hong, S.-N., and Heath, R.~W.
\newblock One-bit sphere decoding for uplink massive {MIMO} systems with
  one-bit {ADC}s.
\newblock \emph{IEEE Transactions on Wireless Communications}, 17\penalty0
  (7):\penalty0 4509--4521, 2018.

\bibitem[Jin et~al.(2020)Jin, Huang, He, Dai, and Wu]{jin2020stochastic}
Jin, R., Huang, Y., He, X., Dai, H., and Wu, T.
\newblock Stochastic-sign {SGD} for federated learning with theoretical
  guarantees.
\newblock \emph{arXiv preprint arXiv:2002.10940}, 2020.

\bibitem[Jin et~al.(2024)Jin, Liu, Huang, He, Wu, and Dai]{jin2024sign}
Jin, R., Liu, Y., Huang, Y., He, X., Wu, T., and Dai, H.
\newblock Sign-based gradient descent with heterogeneous data: Convergence and
  {B}yzantine resilience.
\newblock \emph{IEEE Transactions on Neural Networks and Learning Systems},
  2024.
\newblock {E}arly Access.

\bibitem[Kairouz et~al.(2021)Kairouz, McMahan, Avent, Bellet, Bennis, Bhagoji,
  Bonawitz, Charles, Cormode, Cummings, D’Oliveira, Eichner, Rouayheb, Evans,
  Gardner, Garrett, Gascón, Ghazi, Gibbons, Gruteser, Harchaoui, He, He, Huo,
  Hutchinson, Hsu, Jaggi, Javidi, Joshi, Khodak, Konecný, Korolova,
  Koushanfar, Koyejo, Lepoint, Liu, Mittal, Mohri, Nock, Özgür, Pagh, Qi,
  Ramage, Raskar, Raykova, Song, Song, Stich, Sun, Suresh, Tramèr, Vepakomma,
  Wang, Xiong, Xu, Yang, Yu, Yu, and Zhao]{kairouz2021advances}
Kairouz, P., McMahan, H.~B., Avent, B., Bellet, A., Bennis, M., Bhagoji, A.~N.,
  Bonawitz, K., Charles, Z., Cormode, G., Cummings, R., D’Oliveira, R. G.~L.,
  Eichner, H., Rouayheb, S.~E., Evans, D., Gardner, J., Garrett, Z., Gascón,
  A., Ghazi, B., Gibbons, P.~B., Gruteser, M., Harchaoui, Z., He, C., He, L.,
  Huo, Z., Hutchinson, B., Hsu, J., Jaggi, M., Javidi, T., Joshi, G., Khodak,
  M., Konecný, J., Korolova, A., Koushanfar, F., Koyejo, S., Lepoint, T., Liu,
  Y., Mittal, P., Mohri, M., Nock, R., Özgür, A., Pagh, R., Qi, H., Ramage,
  D., Raskar, R., Raykova, M., Song, D., Song, W., Stich, S.~U., Sun, Z.,
  Suresh, A.~T., Tramèr, F., Vepakomma, P., Wang, J., Xiong, L., Xu, Z., Yang,
  Q., Yu, F.~X., Yu, H., and Zhao, S.
\newblock Advances and open problems in federated learning.
\newblock \emph{Foundations and Trends{\textregistered} in Machine Learning},
  14\penalty0 (1--2):\penalty0 1--210, 2021.

\bibitem[Karimireddy et~al.(2019)Karimireddy, Rebjock, Stich, and
  Jaggi]{karimireddy2019error}
Karimireddy, S.~P., Rebjock, Q., Stich, S., and Jaggi, M.
\newblock Error feedback fixes sign{SGD} and other gradient compression
  schemes.
\newblock In \emph{Proceedings of the 36th International Conference on Machine
  Learning (ICML)}, pp.\  3252--3261, Long Beach, CA, 2019.

\bibitem[Karimireddy et~al.(2021)Karimireddy, He, and
  Jaggi]{karimireddy2021learning}
Karimireddy, S.~P., He, L., and Jaggi, M.
\newblock Learning from history for {B}yzantine robust optimization.
\newblock In \emph{Proceedings of the 38th International Conference on Machine
  Learning (ICML)}, pp.\  5311--5319, Online, 2021.

\bibitem[Kearns \& Saul(1998)Kearns and Saul]{kearns2013large}
Kearns, M. and Saul, L.
\newblock Large deviation methods for approximate probabilistic inference.
\newblock In \emph{Proceedings of the 14th Conference on Uncertainty in
  Artificial Intelligence (UAI)}, Madison, WI, 1998.

\bibitem[Kim et~al.(2019)Kim, Hong, and Lee]{kim2019supervised}
Kim, D., Hong, S.-N., and Lee, N.
\newblock Supervised-learning for multi-hop {MU-MIMO} communications with
  one-bit transceivers.
\newblock \emph{IEEE Journals on Selected Areas in Communications}, 37\penalty0
  (11):\penalty0 2559--2572, 2019.

\bibitem[Kim et~al.(2023{\natexlab{a}})Kim, Lee, and Chung]{kim2023worker}
Kim, D., Lee, J., and Chung, H.~W.
\newblock A worker-task specialization model for crowdsourcing: Efficient
  inference and fundamental limits.
\newblock \emph{IEEE Transactions on Information Theory}, 2023{\natexlab{a}}.
\newblock {E}arly {A}ccess.

\bibitem[Kim et~al.(2023{\natexlab{b}})Kim, Shin, Cassuto, and
  Varshney]{kim2023distributed}
Kim, Y., Shin, J., Cassuto, Y., and Varshney, L.~R.
\newblock Distributed boosting classification over noisy communication
  channels.
\newblock \emph{IEEE Journals on Selected Areas in Communications}, 41\penalty0
  (1):\penalty0 141--154, 2023{\natexlab{b}}.

\bibitem[Krizhevsky \& Hinton(2009)Krizhevsky and
  Hinton]{krizhevsky2009learning}
Krizhevsky, A. and Hinton, G.
\newblock Learning multiple layers of features from tiny images.
\newblock Technical report, University of Toronto, Toronto, Canada, 2009.

\bibitem[Lamport et~al.(1982)Lamport, Shostak, and Pease]{lamport2019byzantine}
Lamport, L., Shostak, R., and Pease, M.
\newblock The {B}yzantine generals problem.
\newblock \emph{ACM Transactions on Programming Languages and Systems},
  4\penalty0 (3):\penalty0 382--401, 1982.

\bibitem[LeCun et~al.(1998)LeCun, Bottou, Bengio, and
  Haffner]{lecun1998gradient}
LeCun, Y., Bottou, L., Bengio, Y., and Haffner, P.
\newblock Gradient-based learning applied to document recognition.
\newblock \emph{Proceedings of the IEEE}, 86\penalty0 (11):\penalty0
  2278--2324, 1998.

\bibitem[Li \& Yu(2014)Li and Yu]{li2014error}
Li, H. and Yu, B.
\newblock Error rate bounds and iterative weighted majority voting for
  crowdsourcing.
\newblock \emph{arXiv preprint arXiv:1411.4086}, 2014.

\bibitem[Li \& Hoefler(2022)Li and Hoefler]{li2022near}
Li, S. and Hoefler, T.
\newblock Near-optimal sparse allreduce for distributed deep learning.
\newblock In \emph{Proceedings of the 27th ACM SIGPLAN Symposium on Principles
  and Practice of Parallel Programming (PPoPP)}, pp.\  135--149, Seoul, South
  Korea, 2022.

\bibitem[Li \& Li(2023)Li and Li]{li2023analysis}
Li, X. and Li, P.
\newblock Analysis of error feedback in federated non-convex optimization with
  biased compression: Fast convergence and partial participation.
\newblock In \emph{Proceedings of the 40th International Conference on Machine
  Learning (ICML)}, pp.\  19638--19688, Honolulu, HI, 2023.

\bibitem[Li et~al.(2019)Li, Huang, Yang, Wang, and Zhang]{li2019convergence}
Li, X., Huang, K., Yang, W., Wang, S., and Zhang, Z.
\newblock On the convergence of fedavg on non-iid data.
\newblock \emph{arXiv preprint arXiv:1907.02189}, 2019.

\bibitem[Li et~al.(2023)Li, Lin, Shang, and Wu]{li2023revisiting}
Li, Z., Lin, T., Shang, X., and Wu, C.
\newblock Revisiting weighted aggregation in federated learning with neural
  networks.
\newblock In \emph{Proceedings of the 40th International Conference on Machine
  Learning (ICML)}, pp.\  19767--19788, Honolulu, HI, 2023.

\bibitem[Lyu et~al.(2020)Lyu, Yu, and Yang]{lyu2020threats}
Lyu, L., Yu, H., and Yang, Q.
\newblock Threats to federated learning: A survey.
\newblock \emph{arXiv preprint arXiv:2003.02133}, 2020.

\bibitem[Park \& Lee(2023{\natexlab{a}})Park and Lee]{park2023s}
Park, C. and Lee, N.
\newblock {$\mathsf{S}^3$GD-MV}: {Sparse-SignSGD} with majority vote for
  communication-efficient distributed learning.
\newblock In \emph{Proceedings of IEEE International Symposium on Information
  Theory (ISIT)}, pp.\  2266--2271, Taipei, Taiwan, 2023{\natexlab{a}}.

\bibitem[Park \& Lee(2023{\natexlab{b}})Park and Lee]{park2023sparse}
Park, C. and Lee, N.
\newblock Sparse-{S}ign{SGD} with majority vote for communication-efficient
  distributed learning.
\newblock \emph{arXiv preprint arXiv:2302.07475}, 2023{\natexlab{b}}.

\bibitem[Pillutla et~al.(2022)Pillutla, Kakade, and
  Harchaoui]{pillutla2022robust}
Pillutla, K., Kakade, S.~M., and Harchaoui, Z.
\newblock Robust aggregation for federated learning.
\newblock \emph{IEEE Transactions on Signal Processing}, 70:\penalty0
  1142--1154, 2022.

\bibitem[Rothchild et~al.(2020)Rothchild, Panda, Ullah, Ivkin, Stoica,
  Braverman, Gonzalez, and Arora]{rothchild2020fetchsgd}
Rothchild, D., Panda, A., Ullah, E., Ivkin, N., Stoica, I., Braverman, V.,
  Gonzalez, J., and Arora, R.
\newblock Fetch{SGD}: Communication-efficient federated learning with
  sketching.
\newblock In \emph{Proceedings of the 37th International Conference on Machine
  Learning (ICML)}, pp.\  8253--8265, Online, 2020.

\bibitem[Sattler et~al.(2019)Sattler, Wiedemann, M{\"u}ller, and
  Samek]{sattler2019robust}
Sattler, F., Wiedemann, S., M{\"u}ller, K.-R., and Samek, W.
\newblock Robust and communication-efficient federated learning from non-i.i.d.
  data.
\newblock \emph{IEEE Transactions on Neural Networks and Learning Systems},
  31\penalty0 (9):\penalty0 3400--3413, 2019.

\bibitem[Seide et~al.(2014)Seide, Fu, Droppo, Li, and Yu]{seide20141}
Seide, F., Fu, H., Droppo, J., Li, G., and Yu, D.
\newblock 1-bit stochastic gradient descent and its application to
  data-parallel distributed training of speech {DNN}s.
\newblock In \emph{Proceedings of the 15th Annual Conference of the
  International Speech Communications Association (INTERSPEECH)}, pp.\
  1058--1062, Singapore, 2014.

\bibitem[Sohn et~al.(2020)Sohn, Han, Choi, and Moon]{sohn2020election}
Sohn, J.-y., Han, D.-J., Choi, B., and Moon, J.
\newblock Election coding for distributed learning: Protecting sign{SGD}
  against {B}yzantine attacks.
\newblock In \emph{Advances in Neural Information Processsing Systems
  (NeurIPS)}, volume~33, pp.\  14615--14625, Online, 2020.

\bibitem[Stich et~al.(2018)Stich, Cordonnier, and Jaggi]{stich2018sparsified}
Stich, S.~U., Cordonnier, J.-B., and Jaggi, M.
\newblock Sparsified {SGD} with memory.
\newblock In \emph{Advances in Neural Information Processsing Systems
  (NeurIPS)}, volume~31, pp.\  4448--4459, Montréal, Canada, 2018.

\bibitem[Sun et~al.(2023)Sun, Wang, Li, and Wang]{sun2023momentum}
Sun, T., Wang, Q., Li, D., and Wang, B.
\newblock Momentum ensures convergence of signsgd under weaker assumptions.
\newblock In \emph{Proceedings of the 40th International Conference on Machine
  Learning (ICML)}, pp.\  33077--33099, Honolulu, HI, 2023.

\bibitem[Wang et~al.(2020)Wang, Sreenivasan, Rajput, Vishwakarma, Agarwal,
  Sohn, Lee, and Papailiopoulos]{wang2020attack}
Wang, H., Sreenivasan, K., Rajput, S., Vishwakarma, H., Agarwal, S., Sohn,
  J.-y., Lee, K., and Papailiopoulos, D.
\newblock Attack of the tails: Yes, you really can backdoor federated learning.
\newblock In \emph{Advances in Neural Information Processing Systems
  (NeurIPS)}, volume~33, pp.\  16070--16084, Online, 2020.

\bibitem[Wangni et~al.(2018)Wangni, Wang, Liu, and Zhang]{wangni2018gradient}
Wangni, J., Wang, J., Liu, J., and Zhang, T.
\newblock Gradient sparsification for communication-efficient distributed
  optimization.
\newblock In \emph{Advances in Neural Information Processsing Systems
  (NeurIPS)}, volume~31, pp.\  1299--1309, Montréal, Canada, 2018.

\bibitem[Wen et~al.(2017)Wen, Xu, Yan, Wu, Wang, Chen, and Li]{wen2017terngrad}
Wen, W., Xu, C., Yan, F., Wu, C., Wang, Y., Chen, Y., and Li, H.
\newblock Tern{G}rad: Ternary gradients to reduce communication in distributed
  deep learning.
\newblock In \emph{Advances in Neural Information Processsing Systems (NIPS)},
  volume~30, pp.\  1--13, Long Beach, CA, 2017.

\bibitem[Wu \& Wang(2021)Wu and Wang]{wu2021fast}
Wu, H. and Wang, P.
\newblock Fast-convergent federated learning with adaptive weighting.
\newblock \emph{IEEE Transactions on Cognitive Communications and Networking},
  7\penalty0 (4):\penalty0 1078--1088, 2021.

\bibitem[Xie et~al.(2020)Xie, Koyejo, and Gupta]{xie2020fall}
Xie, C., Koyejo, O., and Gupta, I.
\newblock Fall of empires: Breaking {B}yzantine-tolerant {SGD} by inner product
  manipulation.
\newblock In \emph{Proceedings of the 35th Uncertainty in Artificial
  Intelligence Conference (UAI)}, volume 115, pp.\  261--270, Online, 2020.

\bibitem[Yin et~al.(2018)Yin, Chen, Kannan, and Bartlett]{yin2018byzantine}
Yin, D., Chen, Y., Kannan, R., and Bartlett, P.
\newblock Byzantine-robust distributed learning: Towards optimal statistical
  rates.
\newblock In \emph{Proceedings of the 35th International Conference on Machine
  Learning (ICML)}, pp.\  5650--5659, Stockholm, Sweden, 2018.

\bibitem[Zhao et~al.(2023)Zhao, Zhou, Li, Tang, Wang, Hou, Min, Zhang, Zhang,
  Dong, Du, Yang, Chen, Chen, Jiang, Ren, Li, Tang, Liu, Liu, Nie, and
  Wen]{zhao2023survey}
Zhao, W.~X., Zhou, K., Li, J., Tang, T., Wang, X., Hou, Y., Min, Y., Zhang, B.,
  Zhang, J., Dong, Z., Du, Y., Yang, C., Chen, Y., Chen, Z., Jiang, J., Ren,
  R., Li, Y., Tang, X., Liu, Z., Liu, P., Nie, J.-Y., and Wen, J.-R.
\newblock A survey of large language models.
\newblock \emph{arXiv preprint arXiv:2303.18223}, 2023.

\bibitem[Zheng et~al.(2019)Zheng, Huang, and Kwok]{zheng2019communication}
Zheng, S., Huang, Z., and Kwok, J.
\newblock Communication-efficient distributed blockwise momentum {SGD} with
  error-feedback.
\newblock In \emph{Advances in Neural Information Processsing Systems
  (NeurIPS)}, volume~32, Vancouver, Canada, 2019.

\bibitem[Zinkevich et~al.(2010)Zinkevich, Weimer, Li, and
  Smola]{zinkevich2010parallelized}
Zinkevich, M., Weimer, M., Li, L., and Smola, A.
\newblock Parallelized stochastic gradient descent.
\newblock In \emph{Advances in Neural Information Processsing Systems (NIPS)},
  volume~23, pp.\  2595--2603, Vancouver, Canada, 2010.

\end{thebibliography}
\bibliographystyle{icml2024}

\newpage
\appendix
\onecolumn
\section{Details of Experiment Settings} \label{app:details}

{\bf Datasets.} The real datasets used for image classification model learning are the MNIST \cite{lecun1998gradient}, CIFAR-10, and CIFAR-100 \cite{krizhevsky2009learning} datasets. MNIST dataset consists of total 70,000 gray-scale image samples of handwritten digits from 0 to 9 with size $28 \times 28$, where 60,000 images are training data and the remaining 10,000 images are test data. CIFAR-10 dataset consists of 60,000 $32 \times 32$ color images in 10 classes with 50,000 training images and 10,000 test images. CIFAR-100 is the same with CIFAR-10, except for 100 classes. Some common data augmentation methods are used such as random cropping, horizontal random flipping, and random rotation. The data distribution for workers is assumed to be IID that each worker has the same number of image samples for the entire classes of each dataset.

{\bf Neural networks.} In the training process, we adopt a convolutional neural network (CNN) \cite{lecun1998gradient} model for the MNIST dataset, and a ResNet \cite{he2016deep} model for the CIFAR-10 and CIFAR-100 datasets. The employed CNN architecture contains $N = 5 \times 10^5$ parameters, consisting of two convolutional layers each with a $5 \times 5$ kernel, connected with two fully-connected layers. For the ResNet model, we use ResNet-56 model which has 56 layers with parameters $N = 8.5 \times 10^5$.

{\bf Benchmarks.} To evaluate the robustness against attacks and the communication efficiency of signSGD-FD, we compare the test accuracy with other conventional DL algorithms. We briefly introduce three signSGD-based DL algorithms and the leveraged parameters, which we adopt in the comparison as below:
\begin{itemize}
    \item Election-signSGD \cite{sohn2020election}: Election-signSGD uses two-stage majority voting aggregation -- multiple polling stations for the first stage, and the final decision aggregation for the second stage. The number of polling stations are the same as the number of workers, and the voting workers for each station is determined by the generator matrix $\mathbf{G}$, which is inspired from coding theory. The matrix $\mathbf{G}$ is generated by the random Bernoulli codes with $r=2$, the best setting for $M=15$ in this paper. Notably, the authors assume that the first stage is not affected by any attack.
    
    \item Sto-signSGD \cite{jin2020stochastic}: Sto-signSGD leverages stochastic sign quantization depending on the magnitude of each gradient component. To be resilient to the attacks, this algorithm additionally uses the reputation-based weights in the majority voting aggregation, which looks similar with our algorithm. We utilize the best parameter $b=0.012$ for CIFAR-10 dataset, where $b$ determines the sign quantization probability for the gradient. The main difference from our algorithm is that sto-signSGD assigns the weight of 0 for the suspected workers, while signSGD-FD can assign the negative weights to those workers.

    \item Noisy signSGD \cite{chen2020distributed}: Noisy signSGD injects random Gaussian noise to the computed local gradients to mitigate the effect of the adversarial attacks. The original goal of this algorithm is to address the heterogeneity of the data distribution by decreasing the gap between mean and median of each gradient component. However, since it is well known that median-based algorithms are attack resistant, we add this Noisy signSGD algorithm to the comparison. We use the standard deviation parameter $b = 10^{-3}$.
\end{itemize}

In the communication costs comparison, we use distributed SGD, Top-K SGD optimizers for the baseline. Considering the robustness against attacks, we apply Multi-Krum algorithm to the above baseline algorithms. We elucidate the details for each algorithm as below:

\begin{itemize}
    \item Distributed SGD \cite{zinkevich2010parallelized}: Distributed SGD leverages 32-bit full-precision information for each gradient component. The aggregation rule is just computing the average of the workers' local gradients.

    \item Top-K SGD \cite{stich2018sparsified}: Top-K SGD selects only the largest $K$ gradient components in magnitude to update its model. As above, the locally computed gradients are averaged in the aggregation process, and the optimizer updates the model by using 32-bit gradient information. In the comparison, we select only 10\% of gradient component, i.e., $\frac{K}{N} = 0.1$.

    \item Multi-Krum \cite{blanchard2017machine}: Multi-Krum is the attack-robust DL algorithm based on SGD optimizers, which trains the model by selecting $K$ workers who do not appear to be affected by the attacks. When there exist $L$ compromised workers in total $M$ workers, the server scores each worker by adding the $\ell$-2 distance between the gradient of corresponding worker and the closest $M-L-2$ gradients of other workers. Then, the server computes the average of gradients by selecting the $K$ lowest-score workers. This can be seen as updating the model with geometric median for workers' gradients.
\end{itemize}

{\bf Communication costs.} In Section \ref{sub:comp_alg}, we compute the total communication cost for each algorithm required to achieve a certain test accuracy. The communication costs per iteration and worker can be calculated by multiplying the number of the shared gradient components per each worker by the bits required to represent each gradient component. We also consider the costs for the uplink communication (from workers to server) and the downlink communication (from server to workers). Based on this, Table I provides the communication costs for each base algorithm.

{\renewcommand{\arraystretch}{1.5}
\begin{table}[ht]
\caption{Total communication costs for each base algorithm.}
\label{sample-table}
\vskip 0.15in
\begin{center}
\begin{sc}
\begin{tabular}{cc}
\toprule
Base algorithms & Total communication costs \\
\midrule
SGD & $\left[ 32N + 32N \right] \times MT$ \\
Top-K SGD & $\left[ \left( 32K + K \log_2 \left( \frac{N}{K} \right) \right) + 32N \right] \times MT$ \\
signSGD-MV & $\left[ N + N \right] \times MT$ \\
\bottomrule
\end{tabular}
\end{sc}
\end{center}
\vskip -0.1in
\end{table}
}

{\bf Hyper-parameters.} For the hyper-parameters that we can tune during the simulations, the number of workers $M$ is fixed to 15 and all workers use the same mini-batch size of $B_m=64, \forall m \in [M]$. The learning rate of each algorithm is carefully selected by comparing the converged test accuracy, where the value is $\delta = 10^{-3}$ and $10^{-1}$ for signSGD-based optimizers and SGD-based optimizers, respectively. To stabilize the learning, we do not actively utilize momentum and weight decay.

\section{Initial Phase Aggregation Design} \label{sub:init_dec}

\begin{figure}[ht]
\vskip 0.2in
\begin{center}
\subfigure[Varying initial decoder type]{%
    \includegraphics[width=0.49\columnwidth]{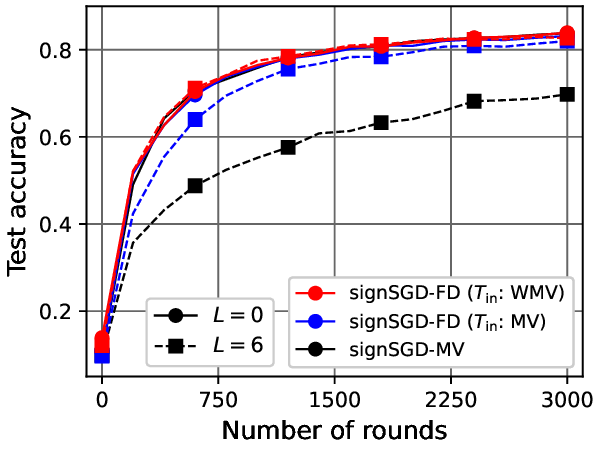}}
\hfill
\subfigure[Varying $T_\mathsf{in}$]{%
    \includegraphics[width=0.49\columnwidth]{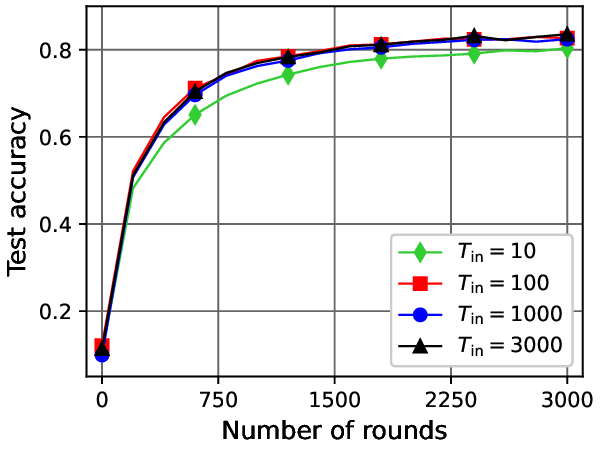}}
\caption{Test accuracy comparison according to the initial phase aggregation of signSGD-FD.}
\label{fig:3} 
\end{center}
\end{figure}

As we observe in Section \ref{sub:anlys_noattack}, we should design the cross-over probability estimation process delicately. To find a better estimation algorithm, we evaluate the test accuracy of signSGD-FV by changing the learning process in the initial phase, which is described in Figure \ref{fig:3}. Here, the number of workers $M$ is 15, and 6 workers are attacked by the SIA method, i.e., $L=6$. The initial aggregation method of signSGD-FD is to perform the WMV aggregation by considering all gradient coordinates in \eqref{eqn:est_prob_init}, but the easiest way is to not use the weights as with the MV aggregation. We compare these two aggregation methods through the test accuracy results for CIFAR-10 image classification task. From the results in Figure \ref{fig:3}-(a), the original signSGD-FD using WMV aggregation has negligible degradation due to attacks, but the accuracy deterioration begins to emerge as we employ the MV aggregation. This is expected to result in inaccurate weight estimation in the initial phase because the majority voting is greatly affected by the attack of $r=1$. The effect of the initial phase duration can be seen in Figure \ref{fig:3}-(b), and this shows us that $T_\mathsf{in}$ does not affect significantly unless the duration is not too short to collect the error samples, such as $T_\mathsf{in} = 10$. Therefore, we can summarize that the initial phase aggregation design should be considered carefully, and the proposed signSGD-FD performs well in the presence of attackers. For ease of implementation, we set the initial period $T_\mathsf{in}$ to 50, 100, 500 for the MNIST, CIFAR-10, and CIFAR-100 datasets, respectively.

\section{Proof of Theorem \ref{thm:1}}
\begin{proof}
    We commence the proof by leveraging Assumption \ref{ass:2} to calculate the upper bound of the loss reduction of $f^{t+1} - f^{t}$. Using the update rule of signSGD-based algorithms in \eqref{eqn:WMV_rule}, the upper bound can be derived as 
    \begin{align}
        f^{t+1} - f^t &\le \left\langle \bar{\mathbf{g}}^t, \mathbf{x}^{t+1} - \mathbf{x}^t \right\rangle + \sum_{n=1}^N \frac{L_n}{2} \left( x_n^{t+1} - x_n^t \right)^2 \nonumber \\
        &= \sum_{n=1}^N \left[ - \bar{g}_n^t \cdot \delta \hat{U}_n^t + \frac{L_n}{2} \left( - \delta \hat{U}_n^t \right)^2 \right] \nonumber \\
        &= - \delta \sum_{n=1}^N \bar{g}_n^t \hat{U}_n^t + \frac{1}{2} \delta^2 \lVert \mathbf{L} \rVert_1 \nonumber \\
        &= - \delta \lVert \bar{\mathbf{g}}^t \rVert_1 + \frac{\delta^2}{2} \lVert \mathbf{L} \rVert_1 + 2\delta \sum_{n=1}^N \left| \bar{g}_n^t \right| \mathbf{1}_{\left[ U_n^t \ne \hat{U}_n^t \right]}. \label{eqn:rate_MV1}
    \end{align}
    By taking expectation according to the randomness of $\hat{U}_n^t$, $f^{t+1} - f^t$ conditioned by $\mathbf{x}^t$ can be upper bounded by
    \begin{align}
        \mathbb{E} \left[ \left. f^{t+1} - f^t \right| \mathbf{x}^t \right] 
        & \le - \delta \lVert \bar{\mathbf{g}}^t \rVert_1 + \frac{\delta^2}{2} \lVert \mathbf{L} \rVert_1 + 2 \delta \sum_{n=1}^N \left| \bar{g}_n^t \right| \mathbb{P} \left[ U_n^t \ne \hat{U}_n^t \right] \nonumber \\
        & \le - \delta \lVert \bar{\mathbf{g}}^t \rVert_1 + \frac{\delta^2}{2} \lVert \mathbf{L} \rVert_1 + 2 \delta P_{\sf E}^{\sf max} \sum_{n=1}^N \left| \bar{g}_n^t \right| \nonumber \\
        &= - \delta \left( 1 - 2 P_{\sf E}^{\sf max}
         \right) \lVert \bar{\mathbf{g}}^t \rVert_1 + \frac{\delta^2}{2} \lVert \mathbf{L} \rVert_1. \label{eqn:rate_MV2}
    \end{align}
    Next, we take the expectation over $\mathbf{x}^t$, and apply a telescoping sum over the iterations, which provides
    \begin{align}
        f^1 - f^\star & \ge f^1 - \mathbb{E} \left[ f^T \right] \nonumber \\
        &= \mathbb{E} \left[ \sum_{t=1}^{T} f^t - f^{t+1} \right] \nonumber \\
        & \ge \mathbb{E} \left[ \sum_{t=1}^{T} \left\{ \delta \left( 1 - 2 P_{\sf E}^{\sf max} \right) \lVert \bar{\mathbf{g}}^t \rVert_1 - \frac{\delta^2}{2} \lVert \mathbf{L} \rVert_1 \right\} \right], \nonumber \\
        &= \delta \left( 1 - 2 P_{\sf E}^{\sf max} \right) \mathbb{E} \left[ \sum_{t=1}^{T} \lVert \bar{\mathbf{g}}^t \rVert_1 \right] -\frac{\delta^2 T}{2} \lVert \mathbf{L} \rVert_1, \label{eqn:rate_MV7}
    \end{align}
    where the last equality holds when $\delta$ is fixed according to the training round $t \in [T]$. Consequently, by plugging the learning rate $\delta = \sqrt{\frac{2 \left( f^1 - f^\star \right)}{T \lVert \mathbf{L} \rVert_1}}$ into \eqref{eqn:rate_MV7}, we obtain 
    \begin{align}
        \mathbb{E} \left[ \frac{1}{T} \sum_{t=1}^{T} \lVert \bar{\mathbf{g}}^t \rVert_1 \right] & \le \frac{1}{1 - 2 P_{\sf E}^{\sf max}} \left[ \frac{1}{\delta T} \left( f^1 - f^\star \right) + \frac{\delta}{2} \lVert \mathbf{L} \rVert_1 \right] \nonumber \\
        &= \frac{1}{1 - 2 P_{\sf E}^{\sf max}} \sqrt{\frac{2 \left( f^1 - f^\star \right) \lVert \mathbf{L} \rVert_1}{T}}. \label{eqn:rate_MV8}
    \end{align}
    This completes the proof. 
\end{proof}

\section{Proof of Theorem \ref{thm:2}} \label{app:thm:2}
\begin{proof}
We define a binary random variable that indicates the decoding error event, i.e.,  $Z_{m,n}^t = {\bf 1}_{\left[ U_n^t \ne Y_{m,n}^t \right]}$. When employing the FD aggregation with the imperfect LLR weight $\hat{w}_{m,n}^t = \ln \frac{1 - \hat{p}_{m,n}^t}{\hat{p}_{m,n}^t}$ with the uncertainty \eqref{eqn:weight_delta}, decoding failures arise if the cumulative sum of weights assigned to the incorrectly decoding workers exceeds half of the total weight. Using this fact, the decoding error probability is rewritten as

\begin{align}
    \mathbb{P} \left[ U_n^t \ne {\hat U}_{{\sf FD}, n}^t \right]&= \mathbb{P} \left[ \sum_{m=1}^M \hat{w}_{m,n}^t Z_{m,n}^t \ge \frac{1}{2} \sum_{m=1}^M \hat{w}_{m,n}^t \right] \nonumber \\
    &= \mathbb{P} \left[ \sum_{m=1}^M \hat{w}_{m,n}^t \left( Z_{m,n}^t - p_{m,n}^t \right) \ge \sum_{m=1}^M \hat{w}_{m,n}^t \left( \frac{1}{2} - p_{m,n}^t \right) \right] \nonumber \\
    &= \mathbb{P} \left[ \sum_{m=1}^M \hat{w}_{m,n}^t \bar{Z}_{m,n}^t \ge \eta \right], \label{eqn:prob_WMV1}
\end{align}
where $\bar{Z}_{m,n}^t = Z_{m,n}^t - p_{m,n}^t$ and $\eta = \sum_{m=1}^M \hat{w}_{m,n}^t \left( \frac{1}{2} - p_{m,n}^t \right)$. Applying Chernoff bound to \eqref{eqn:prob_WMV1} for $s>0$ yields an upper bound on the error probability, expressed as:
\begin{align}
    \mathbb{P} \left[ \sum_{m=1}^M \hat{w}_{m,n}^t \bar{Z}_{m,n}^t \ge \eta \right] & \le \min_{s>0} e^{-\eta s} \, \mathbb{E} \left[ \exp \left( s \sum_{m=1}^M \hat{w}_{m,n}^t \bar{Z}_{m,n}^t \right) \right] \nonumber \\
    & = \min_{s>0} e^{-\eta s} \prod_{m=1}^M \mathbb{E} \left[ e^{s \hat{w}_{m,n}^t \bar{Z}_{m,n}^t} \right].\label{eqn:prob_WMV2}
\end{align}
Here, we leverage the large deviation bound established in Lemma 1 in \cite{kearns2013large}, which is stated as
\begin{align}
    (1-p) e^{-tp} + p e^{t(1-p)} \le \exp \left( \frac{1 - 2p}{4 \ln \frac{1-p}{p}} t^2 \right),
\end{align}
for all $p \in [0, 1]$ and $|t| < \infty$. Then, we obtain the upper bound of the expectation term in \eqref{eqn:prob_WMV2} as 
\begin{align}
    \mathbb{E} \left[ e^{s \hat{w}_{m,n}^t \bar{Z}_{m,n}^t} \right] &= p_{m,n}^t e^{s \hat{w}_{m,n}^t \left( 1 - p_{m,n}^t \right)} + \left( 1 - p_{m,n}^t \right) e^{-s \hat{w}_{m,n}^t p_{m,n}^t} \nonumber \\
    & \le \exp \left[ \frac{1 - 2 p_{m,n}^t}{4 \ln \frac{1 - p_{m,n}^t}{p_{m,n}^t}} \left( \hat{w}_{m,n}^t \right)^2 s^2 \right] \nonumber \\
    &= \exp \left[ \frac{1}{2} \left( \frac{1}{2} - p_{m,n}^t \right) \frac{\hat{w}_{m,n}^t}{w_{m,n}^t} \hat{w}_{m,n}^t s^2 \right], 
\end{align}
where the last equality follows from $w_{m,n}^t=\ln \frac{1 - p_{m,n}^t}{p_{m,n}^t}$. From the uncertainty of LLR weight in \eqref{eqn:weight_delta}, we can express the upper bound as
\begin{align}
    \mathbb{E} \left[ e^{s \hat{w}_{m,n}^t \bar{Z}_{m,n}^t} \right] \le \exp \left[ \frac{1 + \delta_\mathsf{max}}{2} \left( \frac{1}{2} - p_{m,n}^t \right) \hat{w}_{m,n}^t s^2 \right]. \label{eqn:prob_WMV3}
\end{align}
Invoking \eqref{eqn:prob_WMV3} into \eqref{eqn:prob_WMV2}, and also using $\eta = \sum_{m=1}^M \hat{w}_{m,n}^t \left( \frac{1}{2} - p_{m,n}^t \right)$, the upper bound of the FD decoding error probability becomes
\begin{align}
    \mathbb{P} \left[ U_n^t \ne {\hat U}_{{\sf FD}, n}^t \right] &\le \min_{s>0} e^{-\eta s} \exp \left[ \frac{\left( 1 + \delta_\mathsf{max} \right) s^2}{2} \sum_{m=1}^M \left( \frac{1}{2} - p_{m,n}^t \right) \hat{w}_{m,n}^t \right] \nonumber \\
    & =  \min_{s>0} \exp \left[ \frac{\left( 1 + \delta_\mathsf{max} \right) \eta}{2} s^2 - \eta s \right] \nonumber \\
    & = \exp \left[ - \frac{1}{2 \left( 1 + \delta_\mathsf{max} \right)} \eta \right], \label{eqn:proof_thm4_last3}
\end{align}
where the last equality follows from the fact that $s = \frac{1}{1 + \delta_\mathsf{max}}$ is the minimizer of the optimization problem in \eqref{eqn:proof_thm4_last3}. Now, we derive a lower bound of $\eta$ in terms of $\delta_\mathsf{min}$ as
\begin{align}
    \eta &= \sum_{m=1}^M \hat{w}_{m,n}^t \left( \frac{1}{2} - p_{m,n}^t \right) \nonumber \\
    &= \sum_{m=1}^M \left( \frac{1}{2} - p_{m,n}^t \right) \frac{\hat{w}_{m,n}^t}{w_{m,n}^t} \hat{w}_{m,n}^t \nonumber \\
    &\ge \left( 1 - \delta_\mathsf{min} \right) \sum_{m=1}^M \left( \frac{1}{2} - p_{m,n}^t \right) w_{m,n}^t. \label{eqn:delta_min_bound}
\end{align}
Consequently, by substituting \eqref{eqn:delta_min_bound} to \eqref{eqn:proof_thm4_last3}, the upper bound becomes
\begin{align}
    \mathbb{P} \left[ U_n^t \ne \hat{U}_{\mathsf{FD}, n}^t \right] & \le \exp \left[ - M \left( \frac{1 - \delta_\mathsf{min}}{1 + \delta_\mathsf{max}} \right) \cdot \frac{1}{2M} \sum_{m=1}^M \left( \frac{1}{2} - p_{m,n}^t \right) \ln \frac{1 - p_{m,n}^t}{p_{m,n}^t} \right] \nonumber \\
    &= \exp \left[ -M \left( \frac{1 - \delta_\mathsf{min}}{1 + \delta_\mathsf{max}} \right) \gamma_\mathcal{M}^\mathsf{WMV} \right]. \label{eqn:proof_thm4_last2}
\end{align}
This concludes the proof.

\end{proof}

\section{Proof of Theorem \ref{thm:3}} \label{app:thm:3}
\begin{proof}
Similar to the proof in Appendix \ref{app:thm:2}, we express the decoding error probability of the MV aggregation in terms of a binary random variable $Z_{m,n}^t = \mathbf{1}_{\left[ U_n^t \ne \hat{U}_{\mathsf{MV}, n}^t \right]}$ as
\begin{align} \label{eqn:proof_lem3_1}
    \mathbb{P} \left[ U_n^t \ne \hat{U}_{\mathsf{MV}, n}^t \right] = \mathbb{P} \left[ \sum_{m=1}^M Z_{m,n}^t \ge \frac{M}{2} \right].
\end{align}
By applying Markov's inequality in \eqref{eqn:proof_lem3_1}, the MV decoding error probability is upper bounded as
\begin{align}
    \mathbb{P} \left[ U_n^t \ne \hat{U}_{\mathsf{MV}, n}^t \right] &\le \underset{s>0}{\min} \,\, e^{-\frac{M}{2} s} \, \mathbb{E} \left[ \exp \left( s \sum_{m=1}^M Z_{m,n}^t \right) \right] \nonumber \\
    &= \underset{s>0}{\min} \,\, e^{-\frac{M}{2} s} \prod_{m=1}^M \mathbb{E} \left[ e^{s Z_{m,n}^t} \right].
\end{align}
Using the moment generating function of $Z_{m,n}^t$ which follows Bernoulli distribution with $p_{m,n}^t = \mathbb{P} \left[ Z_{m,n}^t = 1 \right]$, the upper bound of the MV decoding error probability becomes
\begin{align} \label{eqn:proof_lem3_2}
    \mathbb{P} \left[ U_n^t \ne \hat{U}_{\mathsf{MV}, n}^t \right] &\le \underset{s>0}{\min} \,\, e^{-\frac{M}{2} s} \prod_{m=1}^M \left( p_{m,n}^t e^s + 1 - p_{m,n}^t \right) \nonumber \\
    &= \underset{s>0}{\min} \,\, \exp \left[ -\frac{M}{2}s + \sum_{m=1}^M \ln \left( p_{m,n}^t \left( e^s - 1 \right) + 1 \right) \right] \nonumber \\
    &\le \underset{s>0}{\min} \,\, \exp \left[ -\frac{M}{2}s + \sum_{m=1}^M p_{m,n}^t \left( e^s - 1 \right) \right],
\end{align}
where the last inequality comes from the property $x \ge \ln (1 + x)$ when $x \ge 0$. Let us denote the average of workers' cross-over probabilities as $\bar{p}_n^t = \frac{1}{M} \sum_{m=1}^M p_{m,n}^t$. Then, by substituting $s = \ln \frac{1}{2 \bar{p}_n^t}$ which minimizes the upper bound in \eqref{eqn:proof_lem3_2}, the upper bound becomes
\begin{align}
   \mathbb{P} \left[ U_n^t \ne \hat{U}_{\mathsf{MV}, n}^t \right] &\le \exp \left[ -\frac{M}{2} \ln \frac{1}{2 \bar{p}_n^t} + M \bar{p}_n^t \left( \frac{1}{2 \bar{p}_n^t} - 1 \right) \right] \nonumber \\
   &= \exp \left[ -M \left( \bar{p}_n^t - \frac{1}{2} \ln \left( 2e \bar{p}_n^t \right) \right) \right] \nonumber \\
   &= \exp \left( -M \gamma_\mathcal{M}^\mathsf{MV} \right).
\end{align}
This concludes the proof.
\end{proof}

\section{Proof of Theorem \ref{thm:4}}
\begin{proof}
In order to analyze the deterioration of the FD aggregation caused by the stochastic sign flip attacks, we leverage the result of Theorem \ref{thm:2}. For ease of expression, we define a function $g(p) = \frac{1}{2} \left( \frac{1}{2} - p \right) \ln \frac{1-p}{p}$ for $p \in [0, 1]$. Then, the FD decoding error bound is derived as
\begin{align}
    \mathbb{P} \left[ U_n^t \ne \hat{U}_{\mathsf{FD}, n}^t \right] &\le \exp \left[ - \left( \frac{1 - \delta_\mathsf{min}}{1 + \delta_\mathsf{max}} \right) \cdot \left( \sum_{m \in \mathcal{M} \backslash \mathcal{L}} g \left( p_{m,n}^t \right) + \sum_{\ell \in \mathcal{L}} g \left( \tilde{p}_{\ell, n}^t \right) \right) \right] \nonumber \\
    &= \exp \left[ - \left( \frac{1 - \delta_\mathsf{min}}{1 + \delta_\mathsf{max}} \right) \cdot \left( \sum_{m \in \mathcal{M}} g \left( p_{m,n}^t \right) - \sum_{\ell \in \mathcal{L}} g \left( p_{\ell,n}^t \right) + \sum_{\ell \in \mathcal{L}} g \left( \tilde{p}_{\ell, n}^t \right) \right) \right] \nonumber \\
    &= \exp \left[ - \left( \frac{1 - \delta_\mathsf{min}}{1 + \delta_\mathsf{max}} \right) \cdot \left( M \gamma_\mathcal{M}^\mathsf{WMV} - L \gamma_\mathcal{L}^\mathsf{WMV} + \sum_{\ell \in \mathcal{L}} g \left( \tilde{p}_{\ell, n}^t \right) \right) \right],
\end{align}
where the error exponent of WMV decoder without attacks $\gamma_\mathcal{M}^\mathsf{WMV} = \frac{1}{M} \sum_{m \in \mathcal{M}} g \left( p_{m,n}^t \right)$ and $\gamma_\mathcal{L}^\mathsf{WMV} = \frac{1}{L} \sum_{\ell \in \mathcal{L}} g \left( p_{\ell,n}^t \right)$ are exploited. Concentrating on the coefficients of error exponents, the error bound of FD aggregation in the presence of attacks becomes
\begin{align}
    \mathbb{P} \left[ U_n^t \ne \hat{U}_{\mathsf{FD}, n}^t \right] \le \exp \left[ - \left( M - L \right) \left( \frac{1 - \delta_\mathsf{min}}{1 + \delta_\mathsf{max}} \right) \tilde{\gamma}_{\mathcal{M}, \mathcal{L}}^\mathsf{WMV} \right],
\end{align}
where $\tilde{\gamma}_\mathcal{L}^\mathsf{WMV} = \frac{1}{M-L} \left[ M \gamma_\mathcal{M}^\mathsf{WMV} - L \gamma_\mathcal{L}^\mathsf{WMV} + \sum_{\ell \in \mathcal{L}} g \left( \tilde{p}_{\ell, n}^t \right) \right]$ is the modified error exponent. This concludes the proof.
\end{proof}

\section{Proof of Corollary \ref{cor:1}}
\begin{proof}
From the result of Theorem \ref{thm:4}, we modify the $\sum_{\ell \in \mathcal{L}} g \left( \tilde{p}_{\ell, n}^t \right)$ term according to the sign flip probability $r = 1 \text{ and } \frac{1}{2}$. Using the compromised cross-over probability in \eqref{eqn:att_prob}, we can derive the property as
\begin{align}
    \frac{1}{2} - \tilde{p}_{\ell, n}^t &= \frac{1}{2} - \left( p_{\ell, n}^t + r \left( 1 - 2 p_{\ell, n}^t \right) \right) \nonumber \\
    &= \left( 1 - 2r \right) \left( \frac{1}{2} - p_{\ell, n}^t \right) \nonumber \\
    &\triangleq q_{\ell, n}^t.
\end{align}
Then, we can organize the $g \left( \tilde{p}_{\ell, n}^t \right)$ term as
\begin{align}
    g \left( \tilde{p}_{\ell, n}^t \right) &= \frac{1}{2} \left( \frac{1}{2} - \tilde{p}_{\ell, n}^t \right) \ln \frac{1 - \tilde{p}_{\ell, n}^t}{\tilde{p}_{\ell, n}^t} \nonumber \\
    &= \frac{1}{2} q_{\ell, n}^t \ln \frac{\frac{1}{2} + q_{\ell, n}^t}{\frac{1}{2} - q_{\ell, n}^t},
\end{align}
where $-\frac{1}{2} < -\left( \frac{1}{2} - p_{\ell, n}^t \right) \le q_{\ell, n}^t \le \frac{1}{2} - p_{\ell, n}^t < \frac{1}{2}$. It can be easily checked that $h(x) = x \ln \frac{1/2 + x}{1/2 - x}$ is an even function, i.e., $h(x) = h(-x)$. Therefore, we can obtain the results that $0 \le \sum_{\ell \in \mathcal{L}} g \left( \tilde{p}_{\ell, n}^t \right) \le \sum_{\ell \in \mathcal{L}} g \left( p_{\ell, n}^t \right) = L \gamma_\mathcal{L}^\mathsf{WMV}$. The lower bound of the inequality can be achieved when $r=\frac{1}{2}$, and the upper bound also can be achieved in the $r=1$ case. By substituting these results to the error exponent $\tilde{\gamma}_\mathcal{L}^\mathsf{WMV}$ in Theorem \ref{thm:4}, the error bound of FD aggregation becomes
\begin{align}
    \mathbb{P} \left[ U_n^t \ne \hat{U}_{\mathsf{FD}, n}^t \right] \le 
    \begin{cases}
        \exp \left[ -M \left( \frac{1 - \delta_\mathsf{min}}{1 + \delta_\mathsf{max}} \right) \gamma_\mathcal{M}^\mathsf{WMV} \right], & \text{if } r = 1 \\
        \exp \left[ - \left( \frac{1 - \delta_\mathsf{min}}{1 + \delta_\mathsf{max}} \right) \left( M \gamma_\mathcal{M}^\mathsf{WMV} - L \gamma_\mathcal{L}^\mathsf{WMV} \right) \right], & \text{if } r = \frac{1}{2}
    \end{cases}.
\end{align}
Since our distributed learning system considers that all workers employ the same mini-batch sizes, i.e., $B_m = B_{m'}, \forall m \ne m'$, which results in the identical cross-over probabilities $p_{m,n}^t = p_{m', n}^t$ by referring the proof of Theorem 1 in \cite{bernstein2018asignsgd}. Therefore, we can note that the error exponents of worker set $\mathcal{M}$ and $\mathcal{L}$ are the same, i.e., $\gamma_\mathcal{M}^\mathsf{WMV} = \gamma_\mathcal{L}^\mathsf{WMV}$. Consequently, we can obtain the final result of Corollary \ref{cor:1} as
\begin{align}
    P_\mathsf{E}^\mathsf{FD} \le
    \begin{cases}
        \exp \left[ -M \left( \frac{1 - \delta_\mathsf{min}}{1 + \delta_\mathsf{max}} \right) \gamma_\mathcal{M}^\mathsf{WMV} \right], & \text{if } r = 1 \\
        \exp \left[ - \left( M-L \right) \left( \frac{1 - \delta_\mathsf{min}}{1 + \delta_\mathsf{max}} \right) \gamma_\mathcal{M}^\mathsf{WMV} \right], & \text{if } r = \frac{1}{2}
    \end{cases}.
\end{align}
\end{proof}

\section{Proof of Theorem \ref{thm:5}}
\begin{proof}
    Aligning with the proof of Theorem \ref{thm:3} in Appendix \ref{app:thm:3}, we can easily derive the upper bound of MV decoding error probability as
    \begin{align} \label{eqn:proof_thm2_1}
        \mathbb{P} \left[ U_n^t \ne \hat{U}_{\mathsf{MV}, n}^t \right] \le \exp \left[ -M \left( \tilde{p}_n^t - \frac{1}{2} \ln \left( 2e \tilde{p}_n^t \right) \right) \right],
    \end{align}
    where $\tilde{p}_n^t$ is the average of cross-over probabilities for benign workers $m \in \mathcal{M} \backslash \mathcal{L}$ and compromised workers $\ell \in \mathcal{L}$, i.e.,
    \begin{align} \label{eqn:proof_thm2_2}
        \tilde{p}_n^t &= \frac{1}{M} \left[ \sum_{m \in \mathcal{M} \backslash \mathcal{L}} p_{m,n}^t + \sum_{\ell \in \mathcal{L}} \tilde{p}_{\ell, n}^t \right] \nonumber \\
        &= \frac{1}{M} \left[ \sum_{m \in \mathcal{M} \backslash \mathcal{L}} p_{m,n}^t + \sum_{\ell \in \mathcal{L}} \left( p_{\ell, n}^t + r \left( 1 - 2 p_{\ell, n}^t \right) \right) \right] \nonumber \\
        &= \frac{1}{M} \left[ \sum_{m \in \mathcal{M}} p_{m,n}^t + r \sum_{\ell \in \mathcal{L}} \left( 1 - 2p_{\ell, n}^t \right) \right] \nonumber \\
        &= \frac{1}{M} \sum_{m \in \mathcal{M}} p_{m,n}^t + \frac{rL}{M} \left( 1 - \frac{2}{L} \sum_{\ell \in \mathcal{L}} p_{\ell, n}^t \right) \nonumber \\
        &= \bar{p}_n^t + \frac{rL}{M} \left( 1 - 2 \bar{p}_{\mathcal{L}, n}^t \right),
    \end{align}
    where $\bar{p}_{\mathcal{L}, n}^t = \frac{1}{L} \sum_{\ell \in \mathcal{L}} p_{\ell, n}^t$ is the average of true cross-over probabilities for the compromised workers $\ell \in \mathcal{L}$. Using \eqref{eqn:proof_thm2_2}, the exponent of error bound in \eqref{eqn:proof_thm2_1} can be expressed as
    \begin{align}
        M \left[ \tilde{p}_n^t - \frac{1}{2} \ln \left( 2e \tilde{p}_n^t \right) \right] &= M \bar{p}_n^t + rL \left( 1 - 2 \bar{p}_{\mathcal{L}, n}^t \right) - \frac{M}{2} \ln \left( 2e \tilde{p}_n^t \right) \nonumber \\
        &= M \left[ \bar{p}_n^t - \frac{1}{2} \ln \left( 2e \bar{p}_n^t \right) \right] + rL \left( 1 - 2 \bar{p}_{\mathcal{L}, n}^t \right) - \frac{M}{2} \ln \frac{\tilde{p}_n^t}{\bar{p}_n^t} \nonumber \\
        &= M \gamma_\mathcal{M}^\mathsf{MV} + rL \left( 1 - 2 \bar{p}_{\mathcal{L}, n}^t \right) - \frac{M}{2} \ln \frac{\tilde{p}_n^t}{\bar{p}_n^t},
    \end{align}
    where the error exponent of MV decoder without attack $\gamma_\mathcal{M}^\mathsf{MV} = \bar{p}_n^t - \frac{1}{2} \ln \left( 2e \bar{p}_n^t \right)$ is used. Likewise, we can continue the proof by leveraging the error exponent term $\gamma_\mathcal{L}^\mathsf{MV} = \bar{p}_{\mathcal{L}, n}^t - \frac{1}{2} \ln \left( 2e \bar{p}_{\mathcal{L}, n}^t \right)$ as
    \begin{align}
        M \left[ \tilde{p}_n^t - \frac{1}{2} \ln \left( 2e \tilde{p}_n^t \right) \right] &= M \gamma_\mathcal{M}^\mathsf{MV} - 2rL \left( \bar{p}_{\mathcal{L}, n}^t - \frac{1}{2} \ln \left( 2e \bar{p}_{\mathcal{L}, n}^t \right) \right) + rL - rL \ln \left( 2e \bar{p}_{\mathcal{L}, n}^t \right) - \frac{M}{2} \ln \frac{\tilde{p}_n^t}{\bar{p}_n^t} \nonumber \\
        &= M \gamma_\mathcal{M}^\mathsf{MV} - 2rL \gamma_\mathcal{L}^\mathsf{MV} - \frac{M}{2} \ln \frac{\tilde{p}_n^t}{\bar{p}_n^t} - rL \ln \frac{\bar{p}_{\mathcal{L}, n}^t}{1/2} \nonumber \\
        &= M \gamma_\mathcal{M}^\mathsf{MV} - 2rL \gamma_\mathcal{L}^\mathsf{MV} + \epsilon_\mathcal{L} (r),
    \end{align}
    and $\epsilon_\mathcal{L} (r) = - \frac{M}{2} \ln \frac{\tilde{p}_n^t}{\bar{p}_n^t} - rL \ln \frac{\bar{p}_{\mathcal{L}, n}^t}{1/2}$ is a sufficiently small parameter which contains the ratios of the average cross-over probabilities. Consequently, if we only focus on the coefficients of the error exponents $\gamma_\mathcal{M}^\mathsf{MV}$ and $\gamma_\mathcal{L}^\mathsf{MV}$, we can simply express the error bound of the MV decoder affected by the stochastic sign flip attacks with probability $r$ as
    \begin{align}
        \mathbb{P} \left[ U_n^t \ne \hat{U}_{\mathsf{MV}, n}^t \right] &\le \exp \left[ - \left( M \gamma_\mathcal{M}^\mathsf{MV} - 2rL \gamma_\mathcal{L}^\mathsf{MV} + \epsilon_\mathcal{L} (r) \right) \right] \nonumber \\
        &= \exp \left[ - \left( M - 2rL \right) \tilde{\gamma}_{\mathcal{M}, \mathcal{L}}^\mathsf{MV} \right],
    \end{align}
    where $\tilde{\gamma}_{\mathcal{M}, \mathcal{L}}^\mathsf{MV} = \frac{1}{M - 2rL} \left[ M \gamma_\mathcal{M}^\mathsf{MV} - 2rL \gamma_\mathcal{L}^\mathsf{MV} + \epsilon_\mathcal{L} (r) \right]$ is the error exponent of MV decoder in the presence of attacks. This concludes the proof.
\end{proof}


\end{document}